\documentclass[journal,twoside]{IEEEtran}    
\usepackage{xcolor}
\usepackage{cite}
\usepackage{amsmath,amssymb,amsfonts}
\usepackage{graphicx}
\usepackage{textcomp}
\usepackage{arydshln}

\graphicspath{{./}{figs/}{fig/}}
\usepackage{graphicx}
\usepackage[dvips]{epsfig} 
\usepackage{epstopdf}
\usepackage{amsmath, bm}
\usepackage{caption}
\usepackage{enumerate}
\usepackage{booktabs}
\usepackage{setspace}
\usepackage{stfloats}
\usepackage{amsthm}
\usepackage{amsfonts}

\usepackage[colorlinks,
linkcolor=cyan,
anchorcolor=cyan,
citecolor=cyan,
hyperfigures= TRUE,
]{hyperref}

\newtheorem{assumption}{Assumption}
\newtheorem{define}{Definition}
\newtheorem{prop}{Proposition}
\newtheorem{lemma}{Lemma}
\newtheorem{thm}{Theorem}
\newtheorem{example}{Example}
\newtheorem{cor}{Corollary}
\usepackage{url}
\newtheorem{rem}{Remark}
\theoremstyle{remark}
\newtheorem*{pf}{\textit{\textbf{Proof}}}

\makeatletter

\newcommand{\Rmnum}[1]{\expandafter\@slowromancap\romannumeral #1@}
\makeatother

\def\QEDopen{{\setlength{\fboxsep}{0pt}\setlength{\fboxrule}{0.2pt}\fbox{\rule[0pt]{0pt}{1.3ex}\rule[0pt]{1.3ex}{0pt}}}} 
\graphicspath{{./}{figs/}{fig/}}

\usepackage{color}

\definecolor{V}{RGB}{18,159,87}
\definecolor{H}{RGB}{255,0,0} 
\usepackage{subfig}
\usepackage{algorithm,algorithmic}
 \usepackage{float}

\def\BibTeX{{\rm B\kern-.05em{\sc i\kern-.025em b}\kern-.08em
		T\kern-.1667em\lower.7ex\hbox{E}\kern-.125emX}}

\begin{document}
	\title{Thompson Sampling-Based Learning and Control for Unknown Dynamic Systems}
	\author{Kaikai Zheng, Dawei Shi, Yang Shi, Long Wang
	\thanks{K. Zheng and D. Shi are with the MIIT Key Laboratory of Servo Motion System Drive and Control, School of Automation, Beijing Institute of Technology, Beijing 100081, China (e-mail: kaikai.zheng@bit.edu.cn, daweishi@bit.edu.cn).}
	\thanks{Y. Shi is with the Department of Mechanical Engineering, Faculty of Engineering, University of Victoria, Victoria, BC V8N 3P6, Canada (e-mail: yshi@uvic.ca).}
	\thanks{Long Wang is with the Center for Systems and Control, College of Engineering, Peking University, Beijing 100871, China (e-mail: longwang@pku.edu.cn).}
	}
\maketitle
	
\begin{abstract}   
	Thompson sampling (TS) is a Bayesian randomized exploration strategy that samples options (e.g., system parameters or control laws) from the current posterior and then applies the selected option that is optimal for a task, thereby balancing exploration and exploitation; this makes TS effective for active learning-based controller design.
	However, TS relies on finite parametric representations, which limits its applicability to more general spaces, which are more commonly encountered in control system design.
	To address this issue, this work proposes a parameterization method for control law learning using reproducing kernel Hilbert spaces and designs a data-driven active learning control approach. 
	Specifically, the proposed method treats the control law as an element in a function space, allowing the design of control laws without imposing restrictions on the system structure or the form of the controller. 
	A TS framework is proposed in this work to reduce control costs through online exploration and exploitation, and the convergence guarantees are further provided for the learning process. 
	Theoretical analysis shows that the proposed method learns the relationship between control laws and closed-loop performance metrics at an exponential rate, and the upper bound of control regret is also derived. 
	Furthermore, the closed-loop stability of the proposed learning framework is analyzed.
	Numerical experiments on controlling unknown nonlinear systems validate the effectiveness of the proposed method.
	\end{abstract}
	
	\begin{IEEEkeywords}
	Learning-based Control, Thompson Sampling, Active Learning Control
	\end{IEEEkeywords}
	
	\section{Introduction}
	In many practical scenarios, such as robotics \cite{chen2024data}, autonomous vehicles\cite{rokonuzzaman2022human}, and industrial processes \cite{markovsky2023data}, the effectiveness of data-driven control critically depends on the availability of high-quality, informative data that sufficiently captures the system’s operational modes and uncertainties. 
	However, obtaining such representative datasets is challenging in practice due to safety, cost, and operational constraints \cite{dorfler2023data1,dorfler2023data2,nortmann2023direct,miller2024robust}.
	To address this issue, active learning control (ALC) improves control performance by actively designing the data acquisition process or enhances data utilization efficiency by actively exploring the most informative sampling points \cite{bachman2017learning,heirung2018model,fasel2022ensemble}.
	Thus, ALC has become a powerful tool for enhancing control performance in complex and uncertain systems, particularly in applications such as robotics \cite{taylor2021active}, industrial automation \cite{fu2019active}, and autonomous vehicles \cite{zhang2024interactive}.

	Among other data-driven control methods, ALC designs data acquisition and utilization processes to improve learning and control effectiveness.
	Depending on the learning objectives, ALC can be categorized into three types: model-based control, policy-based control, and dual control.
	Model-oriented methods aim to obtain accurate or reliable models \cite{mesbah2018stochastic,capone2020localized,hullermeier2021aleatoric,nguyen2022measure,saviolo2023active}.
	In \cite{8759089}, the controller is chosen by maximizing the Fisher information to actively learn the Koopman operator dynamics of robots. 
	Policy-oriented methods focus on exploring the impact of different strategies (control laws) \cite{menard2021fast,chen2022statistical}.
	A reward-free reinforcement learning method was proposed in \cite{chen2022statistical}, aiming to explore the relationship between different strategies and reward functions.
	In the above two types of methods, control performance during the exploration process is often a secondary consideration or even ignored.
	Dual control, on the other hand, utilizes existing information to not only experiment with the system to learn about its behavior and improve control in the future but also effectively control the system \cite{umenberger2019robust, venkatasubramanian2024sequential,li2024dual}.
	In dual control, exploration is usually conducted under the premise that certain closed-loop properties of the system are guaranteed, such as stability \cite{umenberger2019robust}, robustness \cite{venkatasubramanian2024sequential}, and regret \cite{li2024dual}, etc.

	To balance exploration and exploitation during active sampling, various sampling methods for uncertain scenarios have been proposed, including the Upper Confidence Bound (UCB) \cite{garivier2011upper} and Thompson Sampling (TS) \cite{mellor2014decision, abeille2017thompson,10592825}. In the UCB method, the policy is always chosen as the one corresponding to the upper confidence bound of the reward function \cite{carpentier2011upper}. In the TS method, the reward corresponding to each option is randomly generated according to the posterior distribution, and the option with the highest reward is selected.
	The TS can naturally incorporate prior knowledge, effectively handle nonstationary processes (such as time-varying system dynamics), and has been shown to have stronger adaptability and higher convergence speed compared to UCB \cite{yang2025comparison}. 
	Therefore, in this work, the TS framework is chosen to design the data-driven active learning control strategy.

	Following the principles of ALC and the effectiveness of TS, a number of works have been proposed to design active learning control strategies using TS.
	Leveraging the success of numerous applications, the classical TS method has been explored for stochastic control problems with parametric uncertainty.
	Early works considered optimization problems where the unknown system parameters are selected from a finite discrete set. For example, the work in \cite{7819546} addressed scenarios where unknown parameters were chosen from a finite discrete set, and the control law is designed based on the learned parameters.
	More related studies include applications in learning parameterized Markov decision processes \cite{gopalan2015thompson}, regularized linear optimization problems \cite{abeille2017linear}, channel selection \cite{LIU2024111684}, and multi-agent games and consensus problems \cite{10592825}.
	Later, in \cite{8626048}, the gap between discrete and continuous spaces was bridged, where the unknown parameters were assumed to belong to a continuous real space. 
	Expanding on these advancements, the ability to effectively learn unknown parameters in continuous spaces has made the TS method more widely applicable in the field of control. 
	For instance, the authors of \cite{shirani2022thompson} considered the control problem of stochastic differential systems with diffusion processes and unknown parameters. 
	They not only analyzed the regret of the learning process but also discussed the stability of the controller, thereby providing a comprehensive framework for such systems. 
	Furthermore, a method for efficiently designing linear-quadratic regulators for linear systems was discussed in \cite{pmlr-v178-kargin22a}, which highlights the versatility of TS in addressing classical control problems.
	Beyond the limitations of finite discrete parameter spaces \cite{7819546} and continuous real spaces \cite{8626048}, TS methods have the potential for further development in optimizing controller design within functional spaces. 
	This motivates the design of a user-friendly and easily generalizable ALC method in this work.

	In this work, we consider complex and unknown nonlinear systems with almost no prior information, and design an active learning control strategy based on TS that is transferable and generalizable. 
	Specifically, the controller is treated as an element in a function space. 
	Online data is utilized to update the posterior distribution function of the optimal control law within the function space. 
	The controller is periodically updated online according to the TS strategy, which is performed in a general function space and does not rely on prior knowledge of system structures. 
	Therefore, the proposed method's generalizability and transferability make it easily applicable to different systems. 
	To achieve this goal, however, several challenges need to be addressed. 
	First, unlike optimizing parameters in real spaces, it is more challenging to characterize function spaces and design algorithms for updating the associated probability distribution functions.
	Second, the control process and learning process are coupled in TS-based ALC methods, which adds extra difficulty to the analysis of learning convergence and control performance.
	Finally, compared to the potential optimal control law, the learned control method is influenced by the construction of the function space, the posterior distribution update strategy, and the TS sampling algorithm, making the analysis of control regret more complex.

	The contributions of this work are summarized as follows:
	\begin{enumerate}
	\item A control law learning method is proposed through parameterization of a Hilbert space. 
	By constructing Hilbert spaces and a convex hull structure, the proposed method can effectively represent complex nonlinear control laws, which further provides theoretical support for control law optimization. 
	Furthermore, the upper bound of performance degradation (compared to the potential optimal control law) caused by the proposed parameterization method is analyzed (Theorem~\ref{Thm:Gerr}).
	\item Within the parameterized Hilbert space, a TS-based control law learning method is designed to balance exploration and exploitation. 
	A Bayesian inference method is proposed to update the posterior distribution of the unknown cost function using online sampled data. 
	Based on the obtained posterior distribution, a control law is further selected according to the TS mechanism.
	As a result, we prove that the cost function learned by the proposed method converges exponentially to a neighborhood of the true cost function (Theorem \ref{thm:conv}).
	\item The control performance of the proposed method is also analyzed. 
	The upper bound of control regret between the proposed method and the optimal control law is studied, and the closed-loop stability is also analyzed. 
	Specifically, we discuss the control regret bound (Theorem \ref{thm:regret}) and closed-loop stability (Theorem \ref{thm:mean_square_stability}) with respect to the impact of function space parameterization and the TS mechanism. Our results show that the control regret comprises a constant part due to function space parameterization and an exponentially decaying part resulting from posterior distribution updates. Furthermore, the state of the closed-loop system is mean-square bounded.
	\end{enumerate}
	
	The remainder of this work is organized as follows: Section \ref{sec:Problem Formulation} formulates the considered ALC problem; Section \ref{sec:Preliminaries} introduces some useful definitions and lemmas; Section \ref{sec:MainResults} presents the main results of this work, including the parameterization of the function space, the learning method based on TS, the learning convergence analysis, and the discussion of the control regret; Section \ref{sec:extensions} provides several extensions and discussions, including reward functions with nonstationary distributions, closed-loop stability, and computational complexity; Section \ref{sec:NumericalExperiments} provides numerical experiments to validate the effectiveness of the proposed methods; and Section \ref{sec:Conclusion} concludes this work and discusses potential future works.

	\noindent\textbf{Notation:}~Random variables are denoted by lowercase letters, and their specific realizations are represented by uppercase letters with subscripts. Sample spaces are indicated by letters like $\mathcal{X,Y}$, and the corresponding $\sigma$-field is denoted using $\Sigma$. For example, let ${\bm x}=[x_1,x_2,\ldots,x_n]^{\rm T}$ be the state of a nonlinear system with samples $X_1,~X_2,\ldots$, which take values in the space $\mathcal{X}$ with $\sigma$-field $\Sigma_{\bm x}$. For a random variable ${\bm x}\in\mathcal{X}$, the expectation, variance, and covariance are denoted as $\mathbb{E}[{\bm x}]$, $\mathbb{V}[{\bm x}]$, and $\mathbb{C}[{\bm x}_1,{\bm x}_2],~{\bm x}_1,{\bm x}_2\in\mathcal{X}$, respectively. The inner product of two vectors ${\bm x}$ and ${\bm y}$ is denoted as $\langle{\bm x},{\bm y}\rangle$. The Cartesian product of sets $\mathcal{G}_1,\mathcal{G}_2$ is represented as $\mathcal{G}_1\times \mathcal{G}_2$. Moreover, we define $\underline{\bm x}\otimes\overline{\bm x}:=\{{\bm x}=[{x}_1,\ldots,{x}_n]^{\rm T}|{x}_i\in\{\underline{x}_i,\overline{x}_i\}\}$. In this work, $n$-dimensional real spaces are denoted as $\mathbb{R}^n$, $\mathbb{N}$ is used to represent the set of natural numbers, and $\mathbb{N}_n:=\{1,2,\ldots,n\}$. Let $[a,b]:=\{a,\ldots,b\}$ for $a,b\in\mathbb{N}$ be a shorthand notation for a continuous set of natural numbers. Similarly, let ${\bm x}_{[a,b]}:=\{x_a,\ldots,x_b\}$ be a shorthand notation for a set consisting of items of the vector ${\bm x}=[x_a,\ldots,x_b]^{\rm T}$. The $\mathcal{L}_2$-norm of a function is denoted as $\|\cdot\|_{\mathcal{L}_2}$, and the $\mathcal{L}_2$-norm of a vector-valued function ${\bm g}=[g_1,\ldots,g_m]^{\rm T}$ is defined as $\|{\bm g}\|_{\mathcal{L}_2}^2:=\sum\limits_{i=1}^m \|g_i\|^2_{\mathcal{L}_2}$. Moreover, `a.s.' is used to denote `almost surely'.
	
	\section{Problem Formulation}\label{sec:Problem Formulation}
	Consider a nonlinear dynamic system of the form
	\begin{align}\label{eq:sys}
	{\bm x}(k+1)={\bm f}({\bm x}(k),{\bm u}(k),{\bm v}(k)),
	\end{align}
	where ${\bm x}(k)\in\mathcal{X}\subset\mathbb{R}^n$ represents the state, ${\bm v}(k)\in\mathcal{V}\subset\mathbb{R}^{n_v}$ denotes unknown disturbances, ${\bm u}(k)\in\mathcal{U}\subset\mathbb{R}^m$ is the control input, and ${\bm f}(\cdot)=[f_0(\cdot),\ldots,f_n(\cdot)]^{\rm T}$ are unknown system dynamics. In this work, the control input is generated by a state-feedback controller ${\bm g}(\cdot)$, i.e., ${\bm u}(k)={\bm g}({\bm x}(k))=[g_1({\bm x}(k)),\ldots,g_m({\bm x}(k))]^{\rm T}\in\mathbb{R}^{m}$.
	
	The control objective of this work is to design a controller ${\bm g}(\cdot)$ for the given unknown system \eqref{eq:sys} and a specific task, such that a cost function $J$ is minimized. In practice, the cost can be computed based on the system states and inputs over time instants $\{0,\ldots,K\}$. A widely used example is the quadratic cost function 
	\begin{align}\label{eq:costJ}
		J\!=\!\sum\limits_{k=0}^{K}({\bm x}(k)^{\rm T}{ Q}{\bm x}(k)\!+\!{\bm u}(k)^{\rm T}{ R}{\bm u}(k)),
	\end{align} 
	where ${Q}$ and ${R}$ are positive definite matrices. Another widely used example is the risk-sensitive cost function \cite{whittle2002risk} 
	{\small 
	\begin{align}\label{eq:cost_riskJ}
		J\!=\!\frac{1}{\alpha_{\rm risk}}\!\log\mathbb{E}\!\left[\exp\!\left(\sum\limits_{k=0}^{K}\alpha_{\rm risk}({\bm x}(\!k\!)^{\rm T}{ Q}{\bm x}(\!k\!)\!+\!{\bm u}(\!k\!)^{\rm T}{ R}{\bm u}(\!k\!))\!\right)\!\right]
	\end{align}
	}\hspace{-5pt}
	where $\alpha_{\rm risk} > 0$ quantifies the degree of risk sensitivity. A higher $\alpha_{\rm risk}$ places greater emphasis on minimizing high-cost scenarios, making the controller more risk-averse, while a lower $\alpha_{\rm risk}$ results in a more risk-neutral behavior.

	For the sake of analysis, we provide an alternative perspective to describe the cost function $J$. For a given system and control task, the cost function $J$ can be regarded as a random function with respect to the controller ${\bm g}(\cdot)$. Therefore, with a slight abuse of notation, the cost is expressed as $J({\bm g})$. The controller design problem can then be formulated as solving the optimization problem $\min\limits_{{\bm g}\in\mathcal{G}} J({\bm g})$, where $\mathcal{G}$ is the set of all possible controllers.
	
	Considering that searching for a control law in the infinite function space $\mathcal{G}$ is an impractical task, especially when the system dynamics $f(\cdot)$ are unavailable, this paper proposes a learning method based on an initial control law. This is feasible in engineering practice, as many devices normally operate using control laws obtained through empirical or other design methods. 
	
	For an initial control law $\check{\bm g}$, we aim to extract its structural information to construct a Hilbert space $\check{\mathcal{G}}$ such that $\check{\bm g}\in \check{\mathcal{G}}$. The Hilbert space is chosen because it has good properties that facilitate the exploration of potential control laws to improve the system's control performance. For example, thanks to the inner product structure of the Hilbert space, orthogonal function bases can be used to represent possible control laws; the distance metric in the Hilbert space allows us to discuss the proximity of different control laws, enabling methods such as gradient descent to solve optimization problems. For a cost function $J$, online data is collected to explore potential optimal control laws to minimize $J$, i.e., $\min\limits_{{\bm g}\in\check{\mathcal{G}}} J$. To facilitate the discussion, assumptions about the initial control law and the cost function are introduced as follows:
	\begin{assumption}\label{assum:Xbar}
	There exists a convex set $\bar{\mathcal{X}}$ satisfying:
	\begin{enumerate}
	\item $\mathcal{X}\subseteq\bar{\mathcal{X}}$;
	\item $\forall~\overline{\bm  x},\underline{\bm x}\in\bar{\mathcal{X}}: \overline{\bm x}\otimes\underline{\bm x}\subset\bar{\mathcal{X}}$.
	\end{enumerate}
	\end{assumption}
	
	\begin{rem}
		Assumption \ref{assum:Xbar} introduces a convex envelope $\bar{\mathcal{X}}$ of the feasible state set $\mathcal{X}$ that serves as the design domain for constructing the controller class and for subsequent analysis. In practice, $\bar{\mathcal{X}}$ is chosen by enlarging the known or anticipated operating region to include safety/modeling margins; it does not restrict the actual closed-loop trajectories and can be tuned to balance conservatism and tractability. The condition $\overline{\bm x}\otimes\underline{\bm x}\subset\bar{\mathcal{X}}$ simply encodes that convex combinations of admissible states remain in $\bar{\mathcal{X}}$.
		Specifically, $\bar{\mathcal{X}}$ can be set as the minimal axis-aligned bounding box $\{\bm x:\|\bm x\|_\infty\le \sup_{\bm z\in\mathcal{X}}\|\bm z\|_\infty\}$. Furthermore, when $\mathcal{X}=\mathbb{R}^n$, the choice $\bar{\mathcal{X}}=\mathbb{R}^n$ can be adopted. In this way, it is ensured that $\mathcal{X}\subseteq\bar{\mathcal{X}}$ and that convex combinations of admissible states remain in $\bar{\mathcal{X}}$.
	\end{rem}
	
	\begin{assumption}\label{assum:LipofJ}
	The cost function $J$ is Lipschitz continuous with Lipschitz constant $L_J$, i.e.,
	\begin{align}
	|J({\bm g}_1)-J({\bm g}_2)|\leq L_J\|{\bm g}_1-{\bm g}_2\|_{\mathcal{L}_2},~{\bm g}_1,{\bm g}_2\in\check{\mathcal{G}},
	\end{align}
	and $J({\bm g})$ is bounded as $J({\bm g})\geq {\bar{J}_{\rm min}}, {\bm g}\in\check{\mathcal{G}}$ with ${\bar{J}_{\rm min}}$ being a positive constant.
	\end{assumption}

	To avoid unnecessary computational burden or potential system oscillations caused by frequent updates, a segment-based learning method is adopted. Concretely, system data is divided into different segments, and the data of a segment is used as a whole for learning. The control law during system operation is updated at the beginning of each segment. Specifically, let $K$ be the segment length, and $t$ be used to represent the segment index, then the dynamic characteristics of the system can be expressed as:
	\begin{align}
	{\bm x}_t(k)&=f({\bm x}_t(k-1),{\bm u}_t(k-1),{\bm v}_t(k)), &0\leq k\leq K.
	\end{align}
	The controller used in the $t$-th segment is written as ${\bm g}_t$.
	
	With the aforementioned scenario, this work discusses the following three questions:
	\begin{enumerate}
		\item How to construct and evaluate the function space $\check{\mathcal{G}}$? 
		\item How to update the control law ${\bm g}_t$ to balance exploration and exploitation?
		\item How to evaluate the closed-loop performance of the learning-based control systems?
	\end{enumerate}

	\section{Preliminaries}\label{sec:Preliminaries}
	In this section, some useful notations and definitions are introduced for use in subsequent analysis.
	\subsection{Neighborhood}
	\begin{define}(Hellinger distance \cite{barron1999consistency})
	Let $p_1$ and $p_2$ be probability densities on a measurable space $(\mathcal{P},\Sigma_p)$ w.r.t. a measure $v$, then the Hellinger distance between $p_1$ and $p_2$ is defined as 
		\begin{align}\label{eq:dis_H}
			D_{\rm H}(p_1,p_2)\!=\!\frac{1}{\sqrt{2}}\left[\int_\mathcal{P} \left(\sqrt{p_1(w)}\!-\!\sqrt{p_2(w)}\right)^2\!\!{\rm d}v(w)\right]^{\frac{1}{2}}.
		\end{align}
	\end{define}
	
	\begin{define}(Kullback-Leibler divergence, KL divergence, \cite{barron1999consistency})
	Let $p_1$ and $p_2$ be probability densities on a measurable space $(\mathcal{P},\Sigma_p)$ w.r.t. a measure $v$, then the Kullback-Leibler divergence from $p_1$ to $p_2$ is defined as 
		\begin{align}\label{eq:dis_KL}
			D_{\rm KL}(p_1||p_2)=\int_\mathcal{P} p_1(w)\log\left(\frac{p_1(w)}{p_2(w)}\right){\rm d}v(w).
		\end{align}
	\end{define}

	\begin{define}\label{def:Hnei}(Neighborhood \cite{8626048})
		Consider a probability density function $p$ on the measurable space $(\mathcal{P},\Sigma_p)$ with metric $d$, a neighborhood with center $p_1$ and radius $\delta$ is defined as 
		\begin{align}\label{eq:defHnei}
			\mathcal{B}(p_1,\delta):=\{p\in\mathcal{P}:d(p,p_1)< \delta\},
		\end{align}
		which complementary set is defined as
		\begin{align}
			\mathcal{B}^{\rm c}(p_1,\delta):=\{p\in\mathcal{P}:d(p,p_1)\geq \delta\}.
		\end{align}
	\end{define}

	\begin{rem}
	Neighborhoods are used to represent a small region around an element. They consist of three key components: a given element (center) $p_1$, a metric (distance) $d$, and a radius (size) $\delta$. In this work, unless otherwise specified, the symbol $\mathcal{B}$ represents neighborhoods under either the Hellinger distance or the KL divergence. When a specific metric is indicated, $\mathcal{B}_{\rm H}(\cdot,\cdot)$ denotes neighborhoods under the Hellinger distance, and $\mathcal{B}_{\rm KL}(\cdot,\cdot)$ denotes neighborhoods under the KL divergence. A potential caveat here is that the KL divergence is not a true metric, as it is asymmetric (i.e., $D_{\rm KL}(p_1||p_2)\neq D_{\rm KL}(p_2||p_1)$) and does not satisfy the triangle inequality. However, when the center $p_1$ is specified, Definition \ref{def:Hnei} is still meaningful for the KL divergence, and thus it is also discussed in this work.
	\end{rem}

	\subsection{Thompson Sampling}
	TS is a sampling strategy for selecting among multiple options with uncertain rewards. A simple example of TS is provided as follows. Consider a scenario where $N_{\rm dice}$ biased dice are available; in each round, we can choose one of these dice to roll, and the number on the dice is our reward. Our objective is to maximize the reward. Although these dice may appear identical, differences in their mass distribution lead to unknown reward distributions. Therefore, when making a decision (choosing which dice to roll), we need to simultaneously consider two goals: how to gather more information about each dice's reward distribution (exploration), and how to maximize the reward in the current round (exploitation).

	TS provides an analytically convenient strategy to address such problems. The specific procedure is:
	\begin{itemize}
		\item Step~1: Initialize a probability distribution (such as a uniform distribution) for the reward of each dice;
		\item Step~2: For each of the $N_{\rm dice}$ dice, draw one sample from the current probability distribution of each dice, resulting in $N_{\rm dice}$ reward samples;
		\item Step~3: Select the dice with the highest sampled reward in Step~2 and roll it;
		\item Step~4: Update the posterior distribution of the selected dice based on the observed reward, and repeat from Step~2.
	\end{itemize}
	In contrast to the above scenario, the objective of this work is to select a particular controller from a set of candidate control laws. The selected controller is evaluated through the closed-loop system dynamics. By online active sampling, we aim to identify control laws that yield higher rewards (i.e., lower costs).

	\section{Main Results}\label{sec:MainResults}
	The proposed learning approach is introduced in this section, which is divided into four parts.
	The first part proposes a parameterization method for the space $\mathcal{G}$ and discusses the relationship between the constructed function space and the optimal control law. 
	Then, a data-driven method is designed to explore the optimal control law within the function space. 
	The third subsection discusses the convergence of the learning process.
	The last part analyzes the closed-loop performance of the system during the learning process, characterizing the expected system cost through the regret compared to the optimal control law.
	
	\subsection{Parameterization of Controllers}
	In this work, the control law ${\bm g}(\cdot)$ is chosen from a function space $\{g|g:\mathcal{X}\rightarrow  \mathcal{U}\}$. Before introducing the function space construction method proposed in this work, we would like to point out that there are many potential feasible construction methods. When partial information about the system dynamics is available, the corresponding function space $\check{\mathcal{G}}$ can be naturally obtained. However, in this work, we focus on the case where the system dynamics are unknown. In this case, the function space $\check{\mathcal{G}}$ is constructed using the initial control law $\check{\bm g}(\cdot)$. Specifically, Hilbert spaces $\check{\mathcal{G}}_i, i\in\{1,\ldots m\}$ are constructed based on the initial control law, and convex hulls $\mathcal{G}_i\subset\check{\mathcal{G}}_i$ are used to learn the control law. The proposed method is inspired by the fact that many existing control laws are designed based on empirical or other design methods, which can be used as a starting point for constructing the function space. 

	For $\tilde{\bm x}\in\mathcal{X}$, $\forall {\bm x}\in\mathcal{X}$, and an index set ${\bm w}\subseteq[1,n]$, we denote a vector ${\bm d}=[d_1,\ldots,d_n]=:({\bm x},\tilde{\bm x})_{\bm w}$ as follows:
	\begin{align}\label{eq:fuc}
	d_i=\left\{\begin{array}{ll}
		x_i,~&{\text{if}}~i\in{\bm w},\\
		\tilde{x}_i,~&{\text{if}}~i\notin{\bm w},
	\end{array}\right.
	\end{align}
	where ${\bm w}$ is a subset of the set $[1,n]=\{1,\ldots,n\}$, including both the empty set and the full set. Therefore, there are $2^n$ possible ${\bm w}$, which are used to construct the basis functions in the following context.

	Then, a proposition on $({\bm x},\tilde{\bm x})_{\bm w}$ is proposed as follows.
	\begin{prop}\label{prop:Xbar}
	If Assumption \ref{assum:Xbar} holds, $\forall {\bm x},\tilde{\bm x}\in\mathcal{X}$ and $\forall {\bm w}\subseteq[1,n]$, $({\bm x},\tilde{\bm x})_{\bm w}\in\bar{\mathcal{X}}$.
	\end{prop}
	\begin{pf}
	The proof can be  achieved by utilizing Assumption \ref{assum:Xbar} and the definition of $\bar{\mathcal{X}}$, and is therefore omitted here.
	\end{pf}

	Using the notations aforementioned and a given initial control law $\bar{\bm g}(\cdot)=[\bar{ g}_1(\cdot),\ldots,\bar{ g}_m(\cdot)]^{\rm T}$, several sets $\mathcal{G}_1,\ldots,\mathcal{G}_m$ are first introduced as
	\begin{align}\label{eq:calGi}
	\mathcal{G}_i:=\Bigg\{{ g}\Bigg|&{ g}=\sum\limits_{{\bm w}\subseteq[1,n]}\alpha_{\bm w}\Gamma g^{(i)}_{{\bm w},\tilde{\bm x}},~\alpha_{\bm w}\geq0,\notag\\
	&\sum\limits_{{\bm w}\subseteq[1,n]}\alpha_{\bm w}=1,\Bigg\},~i\in\{1,\ldots,m\},
	\end{align}
	where $\Gamma>0$ is a user-defined constant, and $g^{(i)}_{{\bm w},\tilde{\bm x}}$ is a function  that depends on the variables listed in ${\bm w}$ and a constant $\tilde{\bm x}$, which satisfies
	\begin{enumerate}[i)]
	\item $g^{(i)}_{\emptyset,\tilde{\bm x}}=\check{g}_i(\tilde{\bm x}),~g^{(i)}_{{\bm w},\tilde{\bm x}}({\bm x})=\check{g}_i(({\bm x},\tilde{\bm x})_{\bm w})-\sum\limits_{{\bm \varpi }\subsetneq{\bm w}}g^{(i)}_{{\bm \varpi },\tilde{\bm x}}({\bm x})$;
	\item $g^{(i)}_{{\bm w},\tilde{\bm x}}({\bm x})=\sum\limits_{\varpi\subseteq w}(-1)^{|{\bm w}|-|{\bm \varpi}|}\check{g}_i(({\bm x},\tilde{\bm x})_{\bm \varpi})$;\label{eq:gsub}
	\item $g^{(i)}_{{\bm w},\tilde{\bm x}}({\bm x})=0, \text{if}~{\bm x}=\tilde{\bm x}$,
	\end{enumerate}
	with $|{\bm w}|$ representing the number of elements in ${\bm w}$.

	An example of constructing the function space $\mathcal{G}_i$ is provided below.
	\begin{example}
	For an initial controller $\check{g}_i({\bm x})=x_1+x_2+x_1x_2$ and a state $\tilde{\bm x}=[1~2]^{\rm T}$, the index set ${\bm w}$ satisfies 
	\begin{align}
		{\bm w}\in\{{\bm w}_1,{\bm w}_2,{\bm w}_3,{\bm w}_4\}=\{\emptyset,\{1\},\{2\},\{1,2\}\}.
	\end{align}
	Then the basis function corresponding to ${\bm w}_1=\emptyset$ can be obtained as $g^{(i)}_{\emptyset,\tilde{\bm x}}\overset{\rm i)}{=}\check{g}_i(\tilde{\bm x})=5$, based on which the basis functions for ${\bm w}_2$ and ${\bm w}_3$ can be calculated as 
	\begin{align*}
		&g^{(i)}_{{\bm w}_2,\tilde{\bm x}}
	\overset{\rm i)}{=}\check{g}_i(({\bm x},\tilde{\bm x})_{{\bm w}_2})\!-\!g^{(i)}_{\emptyset,\tilde{\bm x}}
	\overset{(\!1\!0\!)}{=}\check{g}_i([x_1~2]^{\rm T})-g^{(i)}_{\emptyset,\tilde{\bm x}}
	=3x_1-3,\\
	&g^{(i)}_{{\bm w}_3,\tilde{\bm x}}
	\overset{\rm i)}{=}\check{g}_i(({\bm x},\tilde{\bm x})_{{\bm w}_3})\!-\!g^{(i)}_{\emptyset,\tilde{\bm x}}
	\overset{(\!1\!0\!)}{=}\check{g}_i([1~x_2]^{\rm T})-g^{(i)}_{\emptyset,\tilde{\bm x}}
	=2x_2-4.
	\end{align*}
	By combining the aforementioned basis functions, we obtain
	\begin{align*}
		&g^{(i)}_{{\bm w}_4,\tilde{\bm x}}\\
		\overset{\rm ii)}{=}&\check{g}_i(({\bm x},\tilde{\bm x})_{{\bm w}_4})\!-\!\check{g}_i(({\bm x},\tilde{\bm x})_{{\bm w}_3})\!-\!\check{g}_i(({\bm x},\tilde{\bm x})_{{\bm w}_2})\!-\!\check{g}_i(({\bm x},\tilde{\bm x})_{{\bm w}_1})\\
		=&x_1x_2-2x_1-x_2+2.
		\end{align*}
	\end{example}

	In the above parameterization method, $\alpha_{\bm w}$ and $\Gamma$ are the weight vector and scaling factor, respectively. Specifically, $\alpha_{\bm w}$ determines the weight of the basis function $g^{(i)}_{{\bm w},\tilde{\bm x}}(\cdot)$ when combining to form a control law. Each vector $\{\alpha_{\bm w}\}$ satisfying the constraints corresponds to a candidate control law in $\mathcal{G}_i$, and the Thompson sampling approach introduced in later sections samples probabilistically from these candidates. The parameter $\Gamma$ is used to adjust the overall amplitude of the basis functions, thereby influencing the output range of the candidate control laws; its value can be designed according to the actual amplitude constraints of the control input.

	The set $\mathcal{G}_i$ can be viewed as the convex hull of the function space $\check{\mathcal{G}}_i$ spanned by a set of basis functions $\{g^{(i)}_{{\bm w},\tilde{\bm x}}\}$. These basis functions are constructed based on the initial control law $\check{g}_i$, with respect to different subsets of the state variables ${\bm w}$ and anchor points $\tilde{\bm x}$. Each $g^{(i)}_{{\bm w},\tilde{\bm x}}$ captures the local variation characteristics of the control law in the corresponding subspace of the state, and they constitute a typical set of bases in the function space. By weighting and combining these basis functions, one can flexibly approximate or represent control strategies in the neighborhood of the initial control law.

	Furthermore, the set $\mathcal{G}$ is defined as the Cartesian product of sets $\mathcal{G}_i,~i\in\{1,\ldots,m\}$ as follows:
	\begin{align}\label{eq:Gsumm}
		\mathcal{G}&:=\mathcal{G}_1\times\mathcal{G}_2\times\ldots\times\mathcal{G}_m.
	\end{align}

	In this work, we assume that the $\mathcal{L}_2$-norm of the initial control law $\check{g}_i$ is bounded, i.e., $\|\check{g}_i\|_{\mathcal{L}_2}\leq M_g$. Consequently, the $\mathcal{L}_2$-norm of the basis functions $g^{(i)}_{{\bm w},\tilde{\bm x}}$ are also bounded, i.e., $\|g^{(i)}_{{\bm w},\tilde{\bm x}}\|_{\mathcal{L}_2}\leq M_g$.

	To analyze the learning error, the following assumption for the optimal controller is adopted.
	\begin{assumption}\label{assum:Gstar}
	For the considered system \eqref{eq:sys}, there exists an optimal controller ${\bm g}^*=[g_1^*,\ldots,g_m^*]^{\rm T},~g_i^*\in\mathcal{G}^*=\left\{g\mid\|g\|_{\mathcal{L}_2}\leq M_g\right\}$ that minimizes the cost function $J({\bm g})$.
	\end{assumption}
	\begin{rem}
	Assumption \ref{assum:Gstar} is used to ensure the existence of the optimal control law. On one hand, in the field of signal processing and control theory, the integral of the square of a signal is often regarded as the energy of the signal. In practice, the bounded energy of most systems makes it natural to consider the $\mathcal{L}_2$ space. Therefore, Assumption \ref{assum:Gstar} only adds a weak premise for the selection of the control law ${\bm g}$. On the other hand, the fact that the square-integrable space is a Hilbert space provides structural advantages for discussing the properties of the optimal solution.
	\end{rem}
	
	We recall that ${\bm g}^*$ is the optimal control law, and $\bar{\bm g}$ is the approximated optimal control law within the function space $\mathcal{G}$ which satisfies $\bar{\bm g}=\arg\min\limits_{{\bm g}\in\mathcal{G}}J$. The constructed function space $\mathcal{G}$ can be quantitatively evaluated by analyzing the regret of $\bar{\bm g}$ with respect to the optimal control law ${\bm g}^*$, which is analyzed as follows.

	\begin{lemma}\label{lem:nfrac}(Lemma 1 in \cite{barron1993universal})
		If ${\bm  g}^*$ is in the closure of the convex hull of a set $G$ in a Hilbert space, with $\|{\bm g}\|_{\mathcal{L}_2}\leq M$ for all ${\bm g}\in G$, there is a $\bar{\bm g}_n$ in the convex hull of $n$ points in $G$ such that 
		\begin{align}
		\|{\bm g}^*-\bar{\bm g}_n\|_{\mathcal{L}_2}^2\leq\frac{M^2-\|{\bm  g}^*\|_{\mathcal{L}_2}}{n}
		\end{align}
		\end{lemma}
		\begin{define}\label{def:rkhs}
		A reproducing kernel Hilbert space (RKHS) $\mathcal{H}({\rm Ker})$ is a Hilbert space of functions on a set $\mathcal{X}$ with a reproducing kernel ${\rm Ker}(\cdot,\cdot)$ that satisfies the following properties:
		\begin{enumerate}[i)]
			\item ${\rm Ker}({\bm x},\cdot)\in\mathcal{H}({\rm Ker})$ for all ${\bm x}\in\mathcal{X}$;
			\item ${\rm Ker}({\bm x}_i,{\bm x}_j)={\rm Ker}({\bm x}_j,{\bm x}_i)$ for all ${\bm x}_i,{\bm x}_j\in\mathcal{X}$;
			\item $\langle g,{\rm Ker}({\bm x},\cdot)\rangle_{\mathcal{H}({\rm Ker})}=g({\bm x})$ for all ${\bm x}\in\mathcal{X}$ and $g\in\mathcal{H}({\rm Ker})$,
		\end{enumerate}
		where $\langle\cdot,\cdot \rangle$ represents the inner product in $\mathcal{H}({\rm Ker})$.
		\end{define}

	With the above tools, we are ready to present our first main result.
	\begin{thm}\label{Thm:Gerr}
	Let Assumptions \ref{assum:Xbar} and \ref{assum:Gstar} hold. Under the proposed parameterization approach \eqref{eq:Gsumm}, the constructed learning set $\mathcal{G}_i$ satisfies: $\exists\bar{\bm g}=\arg\min\limits_{{\bm g}\in\mathcal{G}_i} J$:
	\begin{align}
	\|\bar{g}_i-{g}_i^*\|_{\mathcal{L}_2}\leq\sqrt{\frac{M_g^2-\|{ g}_i^*\|_{\mathcal{L}_2}^2}{2^n}},~\forall i\in\{1,\ldots,m\}.
	\end{align}
	\end{thm}
	\begin{pf}
	The proof of Theorem \ref{Thm:Gerr} can be easily completed by analyzing the properties of the set $\mathcal{G}_i$.
	
	Specifically, according to the definition of $\mathcal{G}_i$ in \ref{def:rkhs}, it can be verified that for $\forall {\bm w}\subseteq[1,n]$, the function $g^{(i)}_{{\bm w},\tilde{\bm x}}$ is in the RKHS $\mathcal{H}(K_{n,{\bm w}}^{(i)})$ with the reproducing kernel $K_{n,{\bm w}}^{(i)}({\bm x},\tilde{\bm x})$. Here, $\mathcal{H}(K_{n,{\bm w}}^{(i)})$ is a Hilbert space of real functions defined on $\mathcal{X}_{\bm w}:=\{{\bm x}_{\bm w}|x\in\mathcal{X}\}$.
	
	According to the definition \eqref{eq:calGi}, functions $g\in\mathcal{G}_i$ can be regarded as linear combinations of base functions $g^{(i)}_{{\bm w},\tilde{\bm x}},~{\bm w}\subseteq[1,n]$. Therefore, for $g\in\mathcal{G}_i$, a RKHS can be obtained as $\mathcal{H}(K_n^{(i)})$ with the reproducing kernel $K_n^{(i)}({\bm x},\tilde{\bm x})$ being
	\begin{align}\label{eq:combK}
	K_d^{(i)}({\bm x},\tilde{\bm x})=\sum\limits_{{\bm w}\subseteq[1,n]}K_{n,{\bm w}}^{(i)}({\bm x}_{\bm w},\tilde{\bm x}_{\bm w}), K_{n,\emptyset }\equiv 1.
	\end{align}
	As a result, the set $\mathcal{G}_i$ can be expressed as the convex hull of $2^n$ points in the RKHS $\mathcal{H}(K_n^{(i)})$.
	Therefore, according to Lemma \ref{lem:nfrac}, there exists a $\bar{\bm g}_i$ in the constructed set $\mathcal{G}_i$ satisfying
	\begin{align}
		\|\bar{g}_i-{g}_i^*\|^2_{\mathcal{L}_2}\leq{\frac{M_g^2-\|{ g}_i^*\|_{\mathcal{L}_2}^2}{2^n}},~\forall i\in\{1,\ldots,m\},
		\end{align}
	which completes the proof of Theorem \ref{Thm:Gerr}.\hfill\QEDopen
	\end{pf} 
	

	\begin{rem}
	Theorem \ref{Thm:Gerr} is an existence result. Specifically, in this work, we construct a function space $\mathcal{G}$ and aim to find a control law $\bar{\bm g}$ in $\mathcal{G}$ to approximate the optimal control law ${\bm g}^*$. By analyzing the properties of the function space $\mathcal{G}$, Theorem~\ref{Thm:Gerr} proves the existence of $\bar{\bm g}$ and the boundedness of regret. The specific method for finding the control law $\bar{\bm g}$ in $\mathcal{G}$ will be described in detail in subsequent sections of this work.
	\end{rem}

	Theorem \ref{Thm:Gerr} addresses the conservatism in approximating the optimal control law within the constructed function space $\mathcal{G}$. This conservatism can be considered from two aspects. First, the construction of $\mathcal{G}$ is determined by the function space spanned by the initial control law, rather than the specific law itself. Thus, even if the initial control law is unstable, as long as the spanned space is rich enough, it is still possible to learn a high-quality control law within $\mathcal{G}$. Second, the conservatism is quantitatively characterized by the distance between $\mathcal{G}$ and the optimal control law ${\bm g}^*$. Specifically, we introduce the reachable error $\Delta_{\mathcal{G}} := \inf_{{\bm g} \in \mathcal{G}} \|{\bm g} - {\bm g}^*\|_{\mathcal{L}_2}$ to capture the approximation error induced by the limitations of the chosen function space. Theorem~\ref{Thm:Gerr} shows that this reachable error is finite under the proposed construction. Moreover, by increasing the dimension of $\mathcal{G}$, the upper bound of the reachable error can be reduced, thereby alleviating the conservatism introduced by the initial function space selection. The reachable error $\Delta_{\mathcal{G}}$ is provided in the following corollary.

	\begin{cor}\label{cor:err}
	Let Assumptions \ref{assum:Xbar} and \ref{assum:Gstar} hold. Under the proposed parameterization approach \eqref{eq:Gsumm}, the reachable error of the constructed function space $\mathcal{G}$ satisfies
	\begin{align}\label{eq:errg}
		\Delta_{\mathcal{G}}\leq\sqrt{\frac{mM_g^2-\sum_{i=1}^m\|{ g}_i^*\|_{\mathcal{L}_2}^2}{2^n}}.
	\end{align}
	\end{cor}
	\begin{pf}
		Corollary \ref{cor:err} can be directly derived from Theorem \ref{Thm:Gerr} and the definition of $\mathcal{L}_2$-norm of vector-valued functions. Thus the proof is omitted here.
	\end{pf}

	\subsection{TS-based Learning Algorithm}
	This subsection introduces an online active sampling method for control laws. Specifically, the proposed method is a policy-oriented ALC approach. Since ${\bm g}(\cdot)$ is not known, evaluating the performance of control laws is challenging. Therefore, this work first designs a TS method to learn the function $J({\bm g})$. At the beginning of each segment, the control law ${\bm u}_t(k)={\bm g}_t({\bm x}_t(k))$ is selected as the one that minimizes the currently learned function, which is denoted as  $J_t({\bm g})$.
	
	In the following, we first introduce the space used to learn the function $J({\bm g})$. Then, an online update method is proposed within a Bayesian framework. To further explain the learning process, pseudocode and corresponding explanations are also provided in this subsection.

	For the considered system \eqref{eq:sys} and a given data segment $\{{\bm x}_t(0),{\bm u}_t(0),\ldots,{\bm x}_t(K_t),{\bm u}_t(K_t)\}$, the control cost can be calculated according to its definition (namely, \eqref{eq:costJ} or \eqref{eq:cost_riskJ}). 
	However, due to the unknown system dynamics $f(\cdot)$ and uncertain noise ${\bm w}$, the cost function $J$ is not available unless the control law is executed. 
	This brings challenges to the controller design. 
	To address this issue, we consider a function space $\bar{\mathcal{J}}$ with $\sigma$-field $\Sigma_{\bar{\mathcal{J}}}$ and measure $v_{\bar{\mathcal{J}}}(\cdot)$ such that the function $\bar{J}({\bm g};{\bm x}_0)\in\bar{\mathcal{J}}$ satisfies the following properties:
	
	\begin{enumerate}
		\item Non-negativity:
				\begin{align}
					\bar{J}({\bm g};{\bm x}_0)>0,~{\bm g}\in\mathcal{G},~{\bm x}_0\in\mathcal{X};
				\end{align}
		\item Probability measure:
				\begin{align}
					\int_{\mathcal{G}}\int_{\mathcal{X}}\bar{J}({\bm g};{\bm x}_0){\rm d}v_{\bar{\mathcal{J}}}({\bm g}){\rm d}{\bm x}_0=1;
				\end{align}
		\item Probability property:
				\begin{align}
					&\forall\Omega\subseteq\mathcal{G},~\forall\Xi\subseteq\mathcal{X}:\notag\\
					&\mathbb{P}({\bm g}\in\Omega,{\bm x}_0\in\Xi)=\int_{\Omega}\int_{\Xi}\bar{J}({\bm g};{\bm x}_0){\rm d}v_{\bar{\mathcal{J}}}({\bm g}){\rm d}{\bm x}_0;
				\end{align}
		\item Proportion:
				\begin{align}\label{eq:prop}
					\bar{J}({\bm g};{\bm x}_0)\propto \frac{1}{J({\bm g},{\bm x}_0)},{\bm g}\in\mathcal{G}, {\bm u}(k)={\bm g}({\bm x}(k)), {\bm x}(0)\!=\!{\bm x}_0,
				\end{align}
				where $J({\bm g},{\bm x}_0)$ denotes the cost function evaluated with control law ${\bm g}$ and initial state ${\bm x}_0$.
		\item Boundedness: $v_{\bar{\mathcal{J}}}(\bar{\mathcal{J}})=\bar{M}_{\bar{\mathcal{J}}},~\bar{J}_{\rm min}\leq\bar{J}({\bm g};{\bm x}_0)\leq\bar{J}_{\rm max}.$
	\end{enumerate}

	Here we note that, through the aforementioned five properties, we establish a space $\bar{\mathcal{J}}$, which serves as a tool for indirectly learning the unknown cost function $J(\cdot)$. Although there are various possibilities for the type of the unknown cost function, under the boundedness assumption \ref{assum:LipofJ}, we can construct a function $\bar{J}({\bm g};{\bm x}_0)$ belonging to the space $\bar{\mathcal{J}}$ and related to the cost function $J(\cdot)$. 
	The inverse in \eqref{eq:prop} is used to convert the cost function into a reward function, which allows higher reward regions to have higher sampling probability according to TS mechanism.
	This facilitates the following learning mechanism design and learning convergence analysis.

	Through the above construction, we obtain a function that characterizes the distribution of the control law ${\bm g}$. Due to property \eqref{eq:prop}, the control law ${\bm g}$ is more likely to be distributed in the region where the cost function $J$ is relatively small. In fact, for a given dynamic characteristic $f(\cdot)$, the distribution of the cost function $J$ can be determined by the control law ${\bm g}$ and the initial state of the system ${\bm x}(0)$. 
	We assume the initial state is consistent across segments, i.e., ${\bm x}_{t_i}(0)={\bm x}_{t_j}(0)$. This can be achieved by designing a reference signal or adjusting segment start times to begin when the system state reaches a predefined value.
	Thus, when it does not cause confusion, $\bar{J}({\bm g};{\bm x})$ will be abbreviated as $\bar{J}({\bm g})$.
	
	In this work, we will indirectly learn the cost function $J({\bm g})$ by learning the constructed function $\bar{J}({\bm g})$. Specifically, let $F_{\rm pdf}(\bar{J})$ be the probability density function (pdf) of the function $\bar{J}$ in the space $\bar{\mathcal{J}}$. The prior distribution is denoted as $F^0_{\rm pdf}(\bar{J})$, and can be updated by the following formula:
	\begin{align}\label{eq:pdfupdate}
	F_{\rm pdf}^{t}(\bar{J})=\frac{R_tF_{\rm pdf}^{0}(\bar{J})}{\int_{\bar{\mathcal{J}}}R_tF_{\rm pdf}^0({\rm d}v_{\mathcal{J}}(\bar{J}))}
	\end{align}
	where $R_t$ is defined as
	\begin{align}
		R_t:=\prod\limits_{i=1}^{t}\frac{\bar{J}({\bm g}_t)}{\bar{J}^*({\bm g}_t)},
	\end{align}
	with $\bar{J}^*({\bm g})$ being 
	\begin{align}\label{eq:defJs}
		\bar{J}^*({\bm g})\propto\frac{1}{J_t({\bm g})}.
	\end{align}
	In the above equation, $J_t({\bm g})$ is the observed cost of the system at $t$-th segment with the control law being ${\bm g}$.
	
	For a subset $\Omega\subseteq\bar{\mathcal{J}}$ and a time instant $t$, the probability $\mathbb{P}(\bar{J}\in\Omega)$ can be calculated using $F_{\rm pdf}^t(\bar{J})$ as 
	\begin{align}\label{eq:JbO}
	\mathbb{P}(\bar{J}\!\in\!\Omega)\!=\!\int_{\Omega}F_{\rm pdf}^t(\bar{J}){\rm d}v_{\mathcal{J}}(\bar{J})\!=\!\frac{\int_\Omega R_tF_{\rm pdf}^{0}({\rm d}v_{\mathcal{J}}(\bar{J}))}{\int_{\bar{\mathcal{J}}}R_tF_{\rm pdf}^0({\rm d}v_{\mathcal{J}}(\bar{J}))}.
	\end{align} 
	
	Based on the above equation, the predictive density at the $t$-th segment can be obtained as the expectation defined as
	\begin{align}\label{eq:JTO}
	\bar{J}^t:=\mathbb{E}\left[\bar{J}\right]:=\int_{\bar{\mathcal{J}}} \bar{J}F_{\rm pdf}^t({\rm d}v_{\mathcal{J}}(\bar{J})),
	\end{align}
	and the restricted and normalized predictive density at the $t$-th segment can be obtained as
	\begin{align}
	\bar{J}^t_\Omega:=\int_{\Omega} \bar{J} \Pi_\Omega^t({\rm d}v_{\mathcal{J}}(\bar{J})),
	\end{align}
	where $\Pi_\Omega^t$ is the posterior distribution restricted and normalized to the set $\Omega$, given by
	\begin{align}
	\Pi_\Omega^t(\cdot)=\frac{\int_{\cdot\cap\Omega}R_tF_{\rm pdf}^0({\rm d}v_{\mathcal{J}}(\bar{J}))}{\int_{\Omega}R_tF_{\rm pdf}^0({\rm d}v_{\mathcal{J}}(\bar{J}))}.
	\end{align}
	

	\begin{algorithm}[H]
	\caption{TS-based ALC.}
	\label{alg:TS}
	\begin{algorithmic}[1]
	\STATE {\textbf{Input}: prior distribution $F_{\rm pdf}^0(\bar{J})$;}\label{line:1}
	\STATE Randomly sample a function $\bar{J}_0$ with initial pdf $F_{\rm pdf}^0(\bar{J})$;
	\STATE Select an initial control law by solving optimization problem ${\bm g}_0=\arg\max\limits_{{\bm g}\in\mathcal{G}} \bar{J}_0({\bm g})$;
	\STATE Apply the control law ${\bm g}_0$ to the system and observe sample data in the initial segment
	\STATE Initialize $R_0=1$;\label{line:initial}
	\FOR{each iteration $t=1,2,\ldots$}
		\STATE Update the posterior distribution:\label{line:updis}
		\begin{align*}
			R_t&=\frac{\bar{J}({\bm g}_{t-1})}{\bar{J}^*({\bm g}_{t-1})}R_{t-1};\\
			F_{\rm pdf}^{t}(\bar{J})&=\frac{R_tF_{\rm pdf}^{0}(\bar{J})}{\int_{\bar{\mathcal{J}}}R_tF_{\rm pdf}^0({\rm d}v_{\mathcal{J}}(\bar{J}))};
			\end{align*}
		\STATE Randomly sample $\bar{J}_t$ according to posterior distribution $F_{\rm pdf}^{t}(\bar{J})$;\label{line:sample}
		\STATE Select the controller ${\bm g}_t$ that maximizes $\bar{J}_t$;\label{line:cont}
		\STATE Apply the controller ${\bm g}_t$ and collect data in the $t$-th segment;\label{line:data}
	\ENDFOR
	\end{algorithmic}
	\end{algorithm}
	
	Based on the distribution function $F_{\rm pdf}^{t}(\bar{J})$, the detailed procedure of the TS adaptive segment-length learning algorithm is illustrated in Algorithm \ref{alg:TS}. The algorithm first sets the prior distribution, samples the performance function, and selects the initial control law. When no other prior knowledge is available, the prior distribution can be chosen as a uniform distribution.
	

	The proposed learning method is an end-to-end control strategy that directly learns the control laws from sampled data, without requiring system model identification and parameter estimation. According to system identification theories \cite{nelles2020nonlinear}, system identification typically assumes that the true system dynamics can be described by a certain class of models. This inevitably limits the design of control laws, as the design is tailored to the model rather than the system itself. The proposed method directly learns the relationship between control laws and costs through random sampling, avoiding assumptions and restrictions on the system model. Thanks to its flexibility, the proposed method can be applied to various types of systems without making assumptions about the system and the distribution of the noise ${\bm v}$.

	\subsection{Convergence Analysis of the Learning Process}
	This subsection discusses the relationship between the function $\bar{J}_t(\cdot)$ obtained under the proposed learning method and the true reward $\bar{J}^*(\cdot)$. 
	To aid our analysis, we write
	\begin{align}\label{eq:Lto}
	{L_t(\Omega)}={\int_{\Omega}R_tF_{\rm pdf}^0({\rm d}v_{\mathcal{J}}(\bar{J}))}
	\end{align}
	for convenience. 
	For KL divergence $D_{KL}$ and Hellinger distance $D_H$, we define a function $T_d(x)$ as
	\begin{align}
		T_d(x)&:=\left\{\!\!\begin{array}{ll}
			\log x, & \text{if}~d=D_{\rm KL},\\
			\sqrt{x}-1, & \text{if}~d=\frac{1}{2}D^2_{\rm H}.
		\end{array}\right.\label{eq:defTd}
	\end{align}
	Furthermore, by combining equations \eqref{eq:Lto} and \eqref{eq:JTO}, we can define a function $T_d^t$ as
	\begin{align}
	T_d^t:=T_d\left(\frac{L_{t}({\mathcal{B}}^{c}(\bar{J}^*,\delta_{\bar{J}}))}{L_{t-1}({\mathcal{B}}^{c}(\bar{J}^*,\delta_{\bar{J}}))}\right),\label{eq:defTdt}
	\end{align}
	where ${\mathcal{B}}(\bar{J}^*,\delta_{\bar{J}})$ is a neighborhood of the predictive density $\bar{J}^*$ with radius $\delta_{\bar{J}}$, which can be defined on eigher KL divergence $d=D_{KL}$ or Hellinger distance $d=\frac{1}{2}D^2_{\rm H}$.
	Then, the non-asymptotic convergence property of the predictive density can be summarized in the following theorem.

	\begin{thm}\label{thm:conv}
	The predictive density $\bar{J}^T$ obtained in \eqref{eq:JTO} satisfies
	\begin{align}\label{eq:convelp}
	\exists \epsilon_L,P_0,\delta_{\bar{J}}\!:\!\mathbb{P}(\bar{J}^T({\bm g})\in{\mathcal{B}}^{c}(\bar{J}^*,\delta_{\bar{J}}))\!<\!P_0\exp(-T\epsilon_L), \text{a.s.},
	\end{align}
	if the following two conditions are satisfied:
	\begin{align}
		&\lim\limits_{t\rightarrow\infty}\inf d(\bar{J}^t_\Omega,\bar{J}^*)>0,~\text{where}~\Omega=\mathcal{B}^{\rm c}(\bar{J}^*,\delta_{\bar{J}})\label{eq:dda0};\\
		&\sum\limits_{t=1}^Tt^{-2}\mathbb{V}(T_d^t)<\infty.\label{eq:VTdt}
	\end{align}
	\end{thm}

	\begin{pf}
		The proof of Theorem \ref{thm:conv} is divided into two steps. In the first step, a martingale process is constructed, and several properties of the martingle are analyzed. In the second step, the proof of Theorem \ref{thm:conv} is completed by using the properties of the constructed martingale process. In the following analysis, the complementary set of the neighborhood ${\mathcal{B}}(\bar{J}^*,\delta_{\bar{J}})$ is denoted as $\Omega={\mathcal{B}}^{c}(\bar{J}^*,\delta_{\bar{J}})$.

		\noindent\textbf{Step~1: martingale analysis.}~
		We consider the $\sigma$-field 
		\begin{align}
		\Sigma_{\bar{{J}}}^t=\sigma(\bar{J}_1,\ldots,\bar{J}_t)=\sigma(\bar{J}({\bm g}_1),\ldots,\bar{J}({\bm g}_t)),
		\end{align}
		which is generated by the control costs of segments $1,\ldots, t$.
		
		According to the definition of $L_{t}$, we have
		\begin{align}
		&\frac{L_{t+1}(\Omega)}{L_t(\Omega)}
		=\frac{\int_{\Omega}R_{t+1}F_{\rm pdf}^{0}({\rm d}{v_{\mathcal{J}}(\bar{J})})}{\int_{\Omega}R_tF_{\rm pdf}^0({\rm d}{v_{\mathcal{J}}(\bar{J})})}\\
		=&\frac{\int_{\Omega}\bar{J}({\bm g}_{t+1})R_{t+1}F_{\rm pdf}^{0}({\rm d}{v_{\mathcal{J}}(\bar{J})})}{\int_{\Omega}\bar{J}({\bm g}_{t+1})R_{t}F_{\rm pdf}^{0}({\rm d}{v_{\mathcal{J}}(\bar{J})})}=\frac{\bar{J}^{t}({\bm g}_{t+1})}{\bar{J}^*({\bm g}_{t+1})}.\label{eq:JbchuJx}
		\end{align}
		
		Then, for the considered $\sigma$-field $\Sigma_{\bar{{J}}}^t$, the following equations can be verified for Hellinger distance $D_{\rm H}(\cdot,\cdot)$:
		\begin{align}
		&\mathbb{E}\left[\left.\sqrt{\frac{L_{t}(\Omega)}{L_{t-1}(\Omega)}}-1\right|\Sigma_{\bar{J}}^t\right]
		=\mathbb{E}\left[\left.\sqrt{\frac{\bar{J}^{t}({\bm g}_{t})}{\bar{J}^*({\bm g}_{t})}}-1\right|\Sigma_{\bar{J}}^t\right]\notag\\
		=&-\frac{1}{2}D_{\rm H}^2(\bar{J}^t(\cdot),\bar{J^*}(\cdot)).\label{eq:EcalH}
		\end{align}
		Similarly, for KL divergence $D_{\rm KL}(\cdot,\cdot)$, we have
		\begin{align}
		&\mathbb{E}\left[\left.\log\frac{L_{t}(\Omega)}{L_{t-1}(\Omega)}\right|\Sigma_{\bar{J}}^t\right]
		=\mathbb{E}\left[\left.\log\frac{\bar{J}^{t}({\bm g}_{t})}{\bar{J}^*({\bm g}_{t})}\right|\Sigma_{\bar{J}}^t\right]\notag\\	
		=&-D_{\rm KL}(\bar{J}^t_{\Omega}(\cdot)||\bar{J^*}(\cdot)).\label{eq:EcalDKL}
		\end{align}
		
		By recalling the definition of $T_d(\cdot)$ in \eqref{eq:defTd}, the following equation can be verified for both Hellinger distance and KL divergence:
		\begin{align}\label{eq:Emd}
		\mathbb{E}\left[\left.T_d\left(\frac{L_{t}(\Omega)}{L_{t-1}(\Omega)}\right)\right|\Sigma_{\bar{{J}}}^t\right]=-d(\bar{J}^t({\bm g}),\bar{J}^*({\bm g})).
		\end{align}
		Then according to the definition of $T_d^t$ defined in \eqref{eq:defTdt}, equation \eqref{eq:Emd} can be equivalently rewritten as the following form for both Hellinger distance and KL divergence:
		\begin{align}
		\mathbb{E}[T_d^t]+d(\bar{J}^t(\cdot),\bar{J}^*(\cdot))=0.\label{eq:sumTDTd}
		\end{align}
		For the stochastic process $M_1,\ldots,M_T$ defined as
		\begin{align}
		M_T=\sum\limits_{t=1}^T\left[T_d^t+d(\bar{J}^t(\cdot),\bar{J}^*(\cdot))\right],\label{eq:MTdef}
		\end{align}
		it can be verified that the following two properties hold:
		\begin{align}
		&\mathbb{E}\left[|M_T|\right]<\infty,\label{eq:MTcond1}\\
		&\mathbb{E}\left[M_{T+1}|M_1,\ldots,M_T\right]=M_T.\label{eq:MTcond2}
		\end{align}
		
		Specifically, from the definition of Hellinger distance \eqref{eq:dis_H} and KL divergence \eqref{eq:dis_KL}, the term $d(\bar{J}^t(\cdot),\bar{J}^*(\cdot))$ is bounded. Condition \eqref{eq:VTdt} ensures that an infinite $T_d^t$ does not exist, i.e., $T_d^t$ is also bounded.
		Thus the inequality \eqref{eq:MTcond1} holds since the sum of multiple bounded items is also bounded.
		Moreover, according to \eqref{eq:MTdef}, we have
		\begin{align}
			M_{T+1}=M_T+T_d^{T+1}+d(\bar{J}^{T+1}(\cdot),\bar{J}^*(\cdot)).\notag
		\end{align}
		Using \eqref{eq:sumTDTd}, the expectation of $M_{T+1}$ can be written as 
		\begin{align}
			\mathbb{E}[M_{T+1}]&=M_T+\mathbb{E}[T_d^{T+1}+d(\bar{J}^{T+1}(\cdot),\bar{J}^*(\cdot))]\notag\\
			&=M_T+0=M_T,\notag
		\end{align}
		which leads to the equation \eqref{eq:MTcond2}.
		
		Thus, the stochastic process $M_T$ is a martingale. 
		The second condition in Theorem \ref{thm:conv} indicates that the variance of the increments of the martingale satisfies the summability condition, i.e., $\sum\limits_{t=1}^Tt^{-2}\mathbb{V}\left\{T_d^t\right\}<\infty$. Therefore, according to the martingale convergence theorem \cite{yau2024principles}, we have
		\begin{align}\label{eq:limMTT}
		\lim\limits_{T\rightarrow\infty}\frac{M_T}{T}=0,~\text{a.s.}
		\end{align}
		Equation \eqref{eq:limMTT} can be equivalently rewritten as
		\begin{align}\label{eq:Td0}
		\lim\limits_{T\rightarrow\infty}\frac{1}{T}\sum\limits_{t=1}^T\!\left[T_d\left(\!\frac{L_{t}(\Omega)}{L_{t-1}(\Omega)}\!\right)\!+\!d(\bar{J}^t(\cdot),\bar{J}^*(\cdot))\!\right]\!=\!0,\text{a.s.}
		\end{align}

		By combining equations \eqref{eq:dda0} and \eqref{eq:Td0}, we have
		\begin{align}\label{eq:supT1T}
		\sup\limits_{T}\frac{1}{T}\sum\limits_{t=1}^TT_d\left(\frac{L_{t}(\Omega)}{L_{t-1}(\Omega)}\right)<0.
		\end{align}
		According to quation \eqref{eq:supT1T}, the following claim can be proved:
		\begin{align}\label{eq:claiminTHM1}
		\exists\epsilon_L(\Omega):~L_T(\Omega)<L_0(\Omega)\exp(-T\epsilon_L), \text{a.s.}
		\end{align}
		The proof of claim \eqref{eq:claiminTHM1} can be performed by analyzing two possible choices of $T_d(\cdot)$ in equation \eqref{eq:defTd}, which are briefly explained as follows.
		
		\noindent\textbf{Case I:~} $T_d(\cdot)=\log(\cdot)$. In this case, equation \eqref{eq:supT1T} can be rewritten as:
		\begin{align}
		\sup\limits_{T}\frac{1}{T}\sum\limits_{t=1}^T\log\left(\frac{L_{t}(\Omega)}{L_{t-1}(\Omega)}\right)<0.
		\end{align}
		By choosing an appropriate $\epsilon_L$ as
		\begin{align}\label{eq:eL1}
		\epsilon_L=-\sup\limits_{T}\frac{1}{T}\sum\limits_{t=1}^T\log\left(\frac{L_{t}(\Omega)}{L_{t-1}(\Omega)}\right),
		\end{align}
		we obtain
		\begin{align}
		\sum\limits_{t=1}^T\log\left(\frac{L_{t}(\Omega)}{L_{t-1}(\Omega)}\right)&<-T\epsilon_L\\
		\Rightarrow L_T(\Omega)&<L_0(\Omega)\exp(-T\epsilon_L),
		\end{align}
		which completes the proof of claim \eqref{eq:claiminTHM1} for Case I.
		
		\noindent\textbf{Case II:~} $T_d(\cdot)=\sqrt{\cdot}-1$. Equation \eqref{eq:supT1T} can be rewritten as:
		\begin{align}
		\sup\limits_{T}\frac{1}{T}\sum\limits_{t=1}^T\left(\sqrt{\frac{L_{t}(\Omega)}{L_{t-1}(\Omega)}}-1\right)<0.
		\end{align}
		In this case, by choosing 
		\begin{align}\label{eq:eL2}
		\epsilon_L=-\sup\limits_{T}\frac{2}{T}\sum\limits_{t=1}^T\left(\sqrt{\frac{L_{t}(\Omega)}{L_{t-1}(\Omega)}}-1\right),
		\end{align}
		we have
		\begin{align}\label{eq:S2TeL}
		\sum\limits_{t=1}^T\left(\sqrt{\frac{L_{t}(\Omega)}{L_{t-1}(\Omega)}}-1\right)<\frac{1}{2}T\epsilon_L.
		\end{align}
		Using the Taylor expansion on the left side of the inequality\eqref{eq:S2TeL}, the following inequality can be obtained:
		\begin{align}
		\log L_{T}(\Omega)<\log L_0-T\epsilon_L,
		\end{align}
		which completes the proof of claim \eqref{eq:claiminTHM1} for Case II by taking the exponential $\exp(\cdot)$ on both sides of the inequality.
		
		\noindent\textbf{Step 2:~Probability analysis.} According to the iterative derivation provided in equation \eqref{eq:pdfupdate} and the probability distribution in \eqref{eq:JbO}, the posterior pdf can be represented as
		\begin{align}
		\mathbb{P}(\bar{J}\in\Omega)&=\int_{\Omega}F_{\rm pdf}^t(\bar{J}){\rm d}{v_{\mathcal{J}}(\bar{J})}\\
		&=\frac{\int_\Omega R_tF_{\rm pdf}^{0}({\rm d}v_{\mathcal{J}}(\bar{J}))}{\int_{\bar{\mathcal{J}}}R_tF_{\rm pdf}^0({\rm d}{v_{\mathcal{J}}(\bar{J})})}\\
		&=\frac{L_{t}(\Omega)}{\int_{\bar{\mathcal{J}}}R_tF_{\rm pdf}^0({\rm d}{v_{\mathcal{J}}(\bar{J})})}.
		\end{align}
		
		Then, according to the claim \eqref{eq:claiminTHM1}, we have
		\begin{align}
		\lim\limits_{t\rightarrow\infty}\mathbb{P}(\bar{J}\in\Omega)=0,~\text{a.s.}
		\end{align}
		As a result, by defining 
		\begin{align}
		P_0=\frac{L_0(\Omega)}{\max\limits_{t}\int_{\bar{\mathcal{J}}}R_tF_{\rm pdf}^t({\rm d}{v_{\mathcal{J}}(\bar{J})})},
		\end{align}
		we obtain 
		\begin{align}
		\mathbb{P}(\bar{J}\in\Omega)<P_0\exp(-N\epsilon_L),~\text{a.s.},
		\end{align}
		which completes the proof of Theorem \ref{thm:conv}. \hfill\QEDopen
	\end{pf}

The first condition \eqref{eq:dda0} in Theorem \ref{thm:conv} requires that outside the neighborhood of $\bar{J}^*$ (specifically, $\Omega(\delta_{\bar{J}}) = \mathcal{B}^{\mathrm{c}}(\bar{J}^*,\delta_{\bar{J}})$), the distance $d(\bar{J}, \bar{J}^*)$ must always be lower bounded by some positive constant. Intuitively, this is essentially an ``identifiability'' or ``positive information gain'' condition: as long as the predictive density $\bar{J}$ is discernibly different from the true density $\bar{J}^*$ (i.e., lies outside the $\delta_{\bar{J}}$-ball), new data can clearly distinguish the current controller from the optimal controller, meaning that each learning step yields strictly positive information gain. In practical control applications, this condition is generally satisfied, because different controllers usually lead to different closed-loop state-input trajectories, which in turn result in distinguishable long-term cost distributions. Moreover, in Theorem \ref{thm:conv}, $\delta_{\bar{J}}$ is a problem-dependent constant rather than a tunable hyperparameter. For a given system and reward function $\bar{J}(\cdot)$, $\delta_{\bar{J}}$ is fixed by the condition \eqref{eq:dda0}.

The second condition \eqref{eq:VTdt} requires that the variance sequence $\{\mathbb{V}(T_d^t)\}$ is square-summable, that is, $\sum_{t=1}^\infty t^{-2}\mathbb{V}(T_d^t)<\infty$. Since each learning segment has a bounded length and the system state, input, and disturbance are all assumed to have bounded second moments (i.e., the working region is bounded and the noise variance is finite), both the cost in each segment and the variance of $T_d^t$ are uniformly bounded by some constant. Essentially, this is a ``light-tailed noise'' requirement on the observation noise, which is met in most practical closed-loop experimental scenarios.

In Theorem \ref{thm:conv}, $\delta_{\bar{J}}$ influences both the neighborhood size and the convergence rate. By definition \eqref{eq:defHnei}, $\delta_{\bar{J}}$ is the radius of the neighborhood $\mathcal{B}(\bar{J}^*,\delta_{\bar{J}})$, so the measure $v_{\bar{\mathcal{J}}}(\mathcal{B}(\bar{J}^*,\delta_{\bar{J}}))$ increases monotonically with $\delta_{\bar{J}}$: a larger $\delta_{\bar{J}}$ means a larger neighborhood and thus worse pointwise accuracy. The convergence rate $\epsilon_L$ of the learning process \eqref{eq:convelp} is determined by the dynamics of the likelihood mass on the complement $\Omega=\mathcal{B}^{\rm c}(\bar{J}^*,\delta_{\bar{J}})$. In general, a larger $\delta_{\bar{J}}$ corresponds to a smaller $\Omega$. Typically, when the current estimate lies in $\Omega$, the empirical reward tends to be smaller relative to $\bar{J}^*$ (as shown in \eqref{eq:JbchuJx}), which leads to a smaller likelihood ratio in the convergence-rate expression and thus a larger $\epsilon_L$. Therefore, a larger $\delta_{\bar{J}}$ typically results in a larger $\epsilon_L$, namely, a faster exponential convergence rate.

		According to Theorem \ref{thm:conv}, it can be claimed that after a finite number of updates, the predictive density $\bar{J}^t$ converges to a neighborhood of the true reward $\bar{J}^*$ with high probability. This is analyzed in the following corollary.
		\begin{cor}\label{cor:time}
		For a constant $P_{\rm fail}\in(0,1)$, if the number of segments $T$ satisfies
		\begin{align}
		T>\frac{1}{\epsilon_L}\log\frac{P_{0}}{P_{\rm fail}},
		\end{align}
		then the probability that the predictive density $\bar{J}^T$ converges to a neighborhood of the true reward $\bar{J}^*$ satisfies
		\begin{align}\label{eq:cor:time}
		\mathbb{P}(\bar{J}^T\in{\mathcal{B}}(\bar{J}^*,\delta_{\bar{J}}))\geq 1-P_{\rm fail}.
		\end{align}
		\end{cor}
		\begin{pf}
			For $T>\frac{1}{\epsilon_L}\log\frac{P_{0}}{P_{\rm fail}}$, we have
			\begin{align}
				P_0\exp(-T\epsilon_L)< P_{\rm fail}.
			\end{align}
			Then according to Theorem \ref{thm:conv}, we obtain
			\begin{align}
				\mathbb{P}(\bar{J}^T\notin{\mathcal{B}}^{\rm c}(\bar{J}^*,\delta_{\bar{J}}))<P_0\exp(-T\epsilon_L)<P_{\rm fail},
			\end{align}
			based on which the conclusion \eqref{eq:cor:time} can be proved. \hfill\QEDopen
		\end{pf}

		\subsection{Analysis of Control Regret}\label{sec:regret}
		In Algorithm \ref{alg:TS}, the control law ${\bm g}_t$ is selected by maximizing the predictive density $\bar{J}^t({\bm g})$. However, the choice of predictive density $\bar{J}^t({\bm g})$ balances exploration and exploitation, i.e., the trade-off between exploring new control laws and exploiting the currently known control laws. The previous subsection focused on the convergence of the learning process, while this subsection focuses on the cost of the control process. Therefore, to distinguish the state under the controller ${\bm g}_t(\cdot)$ from the optimal controller ${\bm g}^*(\cdot)$, we use ${\bm x}^*(k)$ to denote the state under the controller ${\bm g}^*(\cdot)$.

		To facilitate the analysis of controlling regret in the proposed method, let $\bar{J}_t(\cdot)\in\bar{\mathcal{J}}$ be a stochastic function with its pdf denoted as $F^t_{\rm pdf}(\bar{J})$. 
		To avoid confusion, the function chosen in the $t$-th segment is $\bar{J}^t$, which can be regarded as a constant, while $\bar{J}_t(\cdot)$ is a random variable. 
		Therefore, the function $\bar{J}^t$ can be considered as an instance of the random variable $\bar{J}_t(\cdot)$. 
		After the $t$-th segment, the collected sampled data can be used to calculate the cost, which is denoted as $\bar{J}_t({\bm g}_t)$ and is available in the learning process.
		Compared with the unknown optimal reward $\bar{J}^*({\bm g}^*)$, the expected regret over the $t$-th period can be defined as:
		\begin{align}\label{eq:regretdefine}
		R_e({\bm g}_t)=\left|\mathbb{E}\left[\bar{J}^*({\bm g}^*)\right]-\mathbb{E}\left[\bar{J}_t({\bm g}_t)\right]\right|.
		\end{align} 
		The upper bound of this regret function is provided in the following theorem.
		\begin{thm}\label{thm:regret}
		Let Assumptions \ref{assum:Xbar}-\ref{assum:Gstar} hold. With the proposed parameterization approach in \eqref{eq:Gsumm} and the implementation of the TS-based learning in Algorithm \ref{alg:TS}, the regret defined in \eqref{eq:regretdefine} satisfies the following upper bound:
		\begin{align}\label{eq:regretbound}
		&R_e({\bm g}_t)\\
		\leq&\bar{J}_{\rm max}v_{\bar{\mathcal{J}}}(\mathcal{B}(\bar{J}^*,\delta_{\bar{J}}))+\frac{\epsilon_J L_J\sqrt{mM_g^2-\sum_{i=1}^m\|{ g}_i^*\|_{\mathcal{L}_2}^2}}{\sqrt{2^n}J_{\rm min}^2}\notag\\
		&+\left[2\bar{M}_{\bar{\mathcal{J}}}-\bar{J}_{\rm max}v_{\bar{\mathcal{J}}}(\mathcal{B}(\bar{J}^*,\delta_{\bar{J}}))\right]P_0\exp(-t\epsilon_L),~\text{a.s.},\notag
		\end{align}
		where $\epsilon_J>0$ is a constant.
		\end{thm}

	\begin{pf}
	According to the triangle inequality, the control regret considered in \eqref{eq:regretdefine} can be decomposed into the sum of two terms as
	\begin{align}\label{eq:regdecom}
	&R_e({\bm g}_t)\notag\\
	\leq& \left|\mathbb{E}\left[\bar{J}_t({\bm g}_t)\right]-\mathbb{E}\left[\bar{J}^*({\bm g}_t)\right]\right|+\left|\mathbb{E}\left[\bar{J}^*({\bm g}_t)\right]-\mathbb{E}\left[\bar{J}^*({\bm g}^*)\right]\right|,
	\end{align}

	We first discuss the term $\left|\mathbb{E}\left[\bar{J}_t({\bm g}_t)\right]-\mathbb{E}\left[\bar{J}^*({\bm g}_t)\right]\right|$ as follows:
	\begin{align}
	&\left|\mathbb{E}\left[\bar{J}_t({\bm g}_t)\right]-\mathbb{E}\left[\bar{J}^*({\bm g}_t)\right]\right|\\
	=&\left|\int_{\bar{\mathcal{J}}}\bar{J}({\bm g}_t)F^t_{\rm pdf}(\bar{J}){\rm d}v_{\bar{\mathcal{J}}}(\bar{J})-\int_{\bar{\mathcal{J}}}\bar{J}({\bm g}_t)F^*_{\rm pdf}(\bar{J}){\rm d}v_{\bar{\mathcal{J}}}(\bar{J})\right|\notag\\
	\leq&\int_{\bar{\mathcal{J}}}\left|\left[F^t_{\rm pdf}(\bar{J})-F^*_{\rm pdf}(\bar{J})\right]\bar{J}({\bm g}_t)\right|{\rm d}v_{\bar{\mathcal{J}}}(\bar{J})\\
	\leq&\int_{\mathcal{B}(\bar{J}^*,\delta_{\bar{J}})}\left|\left[F^t_{\rm pdf}(\bar{J})-F^*_{\rm pdf}(\bar{J})\right]\bar{J}({\bm g}_t)\right|{\rm d}v_{\bar{\mathcal{J}}}(\bar{J})\label{eq:ftFmF}\\
	&+\int_{\mathcal{B}^{\rm c}(\bar{J}^*,\delta_{\bar{J}})}\left|\left[F^t_{\rm pdf}(\bar{J})-F^*_{\rm pdf}(\bar{J})\right]\bar{J}({\bm g}_t)\right|{\rm d}v_{\bar{\mathcal{J}}}(\bar{J}).
	\end{align}
	Furthermore, the first term in \eqref{eq:ftFmF} can be bounded as
	\begin{align}\label{eq:term11}
	&\int_{\mathcal{B}(\bar{J}^*,\delta_{\bar{J}})}\left|\left[F^t_{\rm pdf}(\bar{J})-F^*_{\rm pdf}(\bar{J})\right]\bar{J}({\bm g}_t)\right|{\rm d}v_{\bar{\mathcal{J}}}(\bar{J})\notag\\
	\leq&v_{\bar{\mathcal{J}}}(\mathcal{B}(\bar{J}^*,\delta_{\bar{J}}))\cdot\max\limits_{{\bm g}_t}\bar{J}({\bm g}_t)\notag\\
	&~\cdot\left[1-\int\limits_{\mathcal{B}(\bar{J}^*,\delta_{\bar{J}})}F^t_{\rm pdf}(\bar{J}){\rm d}v_{\bar{\mathcal{J}}}(\bar{J})\right]\notag\\
	\leq&(1-P_0\exp(-t\epsilon_L))\bar{J}_{\rm max}v_{\bar{\mathcal{J}}}(\mathcal{B}(\bar{J}^*,\delta_{\bar{J}})).
	\end{align}
	
	According to the boundedness of the function space $\bar{\mathcal{J}}$ and Theorem \ref{thm:conv}, the second term in \eqref{eq:ftFmF} can be further bounded as 
	\begin{align}\label{eq:term12}
	&\int_{\mathcal{B}^{\rm c}(\bar{J}^*,\delta_{\bar{J}})}\left|\left[F^t_{\rm pdf}(\bar{J})-F^*_{\rm pdf}(\bar{J})\right]\bar{J}({\bm g}_t)\right|{\rm d}v_{\bar{\mathcal{J}}}(\bar{J})\notag\\
	\leq&2\bar{M}_{\bar{\mathcal{J}}}P_0\exp(-t\epsilon_L).
	\end{align}
	
	By combining equations \eqref{eq:term11} and \eqref{eq:term12}, we have
	\begin{align}\label{eq:boundterm1}
	&\left|\mathbb{E}\left[\bar{J}_t({\bm g}_t)\right]-\mathbb{E}\left[\bar{J}^*({\bm g}_t)\right]\right|\\
	\leq&(1-P_0\exp(-t\epsilon_L))\bar{J}_{\rm max}v_{\bar{\mathcal{J}}}(\mathcal{B}(\bar{J}^*,\delta_{\bar{J}}))+2\bar{M}_{\bar{\mathcal{J}}}P_0\exp(-t\epsilon_L)\notag\\
	\leq&\bar{J}_{\rm max}v_{\bar{\mathcal{J}}}(\mathcal{B}(\bar{J}^*,\delta_{\bar{J}}))\!+\!\left[2\bar{M}_{\bar{\mathcal{J}}}\!-\!\bar{J}_{\rm max}v_{\bar{\mathcal{J}}}(\mathcal{B}(\bar{J}^*,\delta_{\bar{J}}))\right]P_0\exp(-t\epsilon_L).\notag
	\end{align}
	
	Then the upper bound of the second term in equation \eqref{eq:regdecom}, i.e., $\left|\mathbb{E}\left[\bar{J}^*({\bm g}_t)\right]-\mathbb{E}\left[\bar{J}^*({\bm g}^*)\right]\right|$, can be derived as follows:
	\begin{align}\label{eq:intJ}
	&\left|\mathbb{E}\left[\bar{J}^*({\bm g}_t)\right]-\mathbb{E}\left[\bar{J}^*({\bm g}^*)\right]\right|\notag\\
	=&\left|\int_{\bar{\mathcal{J}}}\bar{J}^*({\bm g}_t)F^*_{\rm pdf}(\bar{J}^*){\rm d}v_{\bar{\mathcal{J}}}(\bar{J}^*)\!-\!\int_{\bar{\mathcal{J}}}\bar{J}^*({\bm g}^*)F^*_{\rm pdf}(\bar{J}^*){\rm d}v_{\bar{\mathcal{J}}}(\bar{J}^*)\right|\notag\\
	\leq&\int_{\bar{\mathcal{J}}}\left|\bar{J}^*({\bm g}_t)-\bar{J}^*({\bm g}^*)\right|F^*_{\rm pdf}(\bar{J}^*){\rm d}v_{\bar{\mathcal{J}}}(\bar{J}^*).
	\end{align}
	By recalling equation \eqref{eq:defJs} that $\bar{J}^* \propto \frac{1}{J^*}$ and writing the proportional coefficient as $\epsilon_J$, we have $\bar{J}^*=\frac{\epsilon_J}{J^*}$.
	According to Assumption \ref{assum:LipofJ} that the function $J({\bm g})$ is Lipschitz continuous and bounded, we obtain
	\begin{align}\label{eq:ineqJggt}
	&\left|\bar{J}^*({\bm g}_t)-\bar{J}^*({\bm g}^*)\right|
	=\left|\frac{\epsilon_J}{J^*({\bm g}_t)}-\frac{\epsilon_J}{J^*({\bm g}^*)}\right|\notag\\
	=&\epsilon_J\frac{\left|J^*({\bm g}_t)-J^*({\bm g}^*)\right|}{|J^*({\bm g}_t)||J^*({\bm g}^*)|}
	\leq \frac{\epsilon_J L_J}{J_{\rm min}^2}\|{\bm g}_t-{\bm g}^*\|_{\mathcal{L}_2}.
	\end{align}
	Using the upper bound provided in Theorem \ref{Thm:Gerr}, inequality \eqref{eq:ineqJggt} can be further derived as
	\begin{align}
	&\left|\bar{J}^*({\bm g}_t)-\bar{J}^*({\bm g}^*)\right|\leq\frac{\epsilon_J L_J\sqrt{mM_g^2-\sum_{i=1}^m\|{ g}_i^*\|_{\mathcal{L}_2}^2}}{\sqrt{2^n}J_{\rm min}^2}.
	\end{align}
	Furthermore, the definition of probabilities lead to
	\begin{align}
	\int_{\bar{\mathcal{J}}}F^*_{\rm pdf}(\bar{J}^*){\rm d}v_{\bar{\mathcal{J}}}(\bar{J}^*)=1.
	\end{align}
	Then, the upper bound of the second term in equation \eqref{eq:intJ} can be derived as
	\begin{align}\label{eq:boundterm2}
	&\left|\mathbb{E}\left[\bar{J}^*({\bm g}_t)\right]-\mathbb{E}\left[\bar{J}^*({\bm g}^*)\right]\right|\notag\\
	\leq&\frac{\epsilon_J L_J\sqrt{mM_g^2-\sum_{i=1}^m\|{ g}_i^*\|_{\mathcal{L}_2}^2}}{\sqrt{2^n}J_{\rm min}^2}.
	\end{align}
	Combining inequalities \eqref{eq:boundterm1} and \eqref{eq:boundterm2}, the regret can be bounded as
	\begin{align}
	R_e({\bm g}_t)\leq&\bar{J}_{\rm max}v_{\bar{\mathcal{J}}}(\mathcal{B}(\bar{J}^*,\delta_{\bar{J}}))\notag\\
	&+\left[2\bar{M}_{\bar{\mathcal{J}}}-\bar{J}_{\rm max}v_{\bar{\mathcal{J}}}(\mathcal{B}(\bar{J}^*,\delta_{\bar{J}}))\right]P_0\exp(-t\epsilon_L)\notag\\
	&+\frac{\epsilon_J L_J\sqrt{mM_g^2-\sum_{i=1}^m\|{ g}_i^*\|_{\mathcal{L}_2}^2}}{\sqrt{2^n}J_{\rm min}^2},
	\end{align}
	which completes the proof.\hfill\QEDopen
	\end{pf}

	\begin{rem}\label{rem:contcontexp}
		In Theorem \ref{thm:regret}, the regret bound contains the term $\bar{J}_{\rm max}v_{\bar{\mathcal{J}}}(\mathcal{B}(\bar{J}^*,\delta_{\bar{J}}))$ and the stationary component $\big[2\bar{M}_{\bar{\mathcal{J}}}-\bar{J}_{\rm max}v_{\bar{\mathcal{J}}}(\mathcal{B}(\bar{J}^*,\delta_{\bar{J}}))\big]P_0\exp(-t\epsilon_L)$. An increase in $\delta_{\bar{J}}$ increases the neighborhood measure $v_{\bar{\mathcal{J}}}(\mathcal{B}(\bar{J}^*,\delta_{\bar{J}}))$ and also increases $\epsilon_L$ (see the discussions in Section IV.C). The overall effect on the regret bound is therefore a tradeoff between a larger stationary component and a faster-decaying transient, which depends on the system considered and the time horizon $t$ of the TS-based ALC.
	\end{rem}

	The regret bound in Theorem \ref{thm:regret} describes the expected gap in reward between the control law ${\bm g}_t$ adopted in the $t$-th segment of the proposed method and the potential optimal control law ${\bm g}^*$. In the literature related to TS, the average per period regret over $T$-periods is often discussed, which is defined as $R_e^{T}=\frac{1}{T}\left|\sum_{t=0}^T R_e({\bm g}_t)\right|$. Based on the regret upper bound \eqref{eq:regretbound} in Theorem \ref{thm:regret} and the analysis of each term in Remark \ref{rem:contcontexp}, we can conclude that the average per period regret over $T$-periods for the proposed method satisfies $R_e^{T}=\mathcal{C}+\mathcal{O}(1/T)$, where $\mathcal{C}$ is the constant formed by the first two terms in \eqref{eq:regretbound}, and $\mathcal{O}(1/T)$ reflects the asymptotic property of the third term in \eqref{eq:regretbound}. Compared to existing literature \cite{8626048}, the proposed method not only characterizes the convergence of the learning process but also discusses the impact of function space parameterization and the hyperparameter $\delta_{\bar{J}}$ on the closed-loop regret. 
	Specifically, for sufficiently small $\delta_{\bar{J}}$, the proposed method can converge to the optimal solution ${\bm g}^*$ within the constructed function space, which represents the unavoidable regret compared with that of selecting the control law from an infinite-dimensional function space. The regret can be further reduced by constructing the function space $\mathcal{G}$ more appropriately.
	It is worth noting that a function space $\mathcal{G}$ with more possibilities also leads to increased computational complexity and slower convergence.

\section{Extensions and Closed-Loop Stability Analysis}\label{sec:extensions}
For analytical convenience, some assumptions were made in the previous sections during the discussion of learning convergence and control regret: (1) the system state at the beginning of each segment is the same; (2) the length of each segment is identical; (3) the system dynamics are time-invariant (namely, reward distribution is stationary). In this section, by introducing additional computational mechanisms, Theorem \ref{thm:conv} and Theorem \ref{thm:regret} are extended to more general scenarios, and the closed-loop stability and computational complexity of the proposed method are discussed.

\vspace{-0.5em}
\subsection{Extensions on Information Gain-Based Sampling Stopping Mechanism and Unknown Initial State Distribution}
\subsubsection{Modification on the Sampling Stopping Criterion}
To improve learning efficiency, the segment length $K_t$ is determined adaptively based on the information content of the collected data, with constraints $K_{\min} \leq K_t \leq K_{\max}$. Specifically, an information gain-based stopping criterion is designed as follows.

\begin{define}\label{def:info_content}
For a probability density function $F_{\rm pdf}(\bar{J})$ on the space $\bar{\mathcal{J}}$, the information content (differential entropy) is defined as
\begin{align}\label{eq:info_content}
H(F_{\rm pdf}) := -\int_{\bar{\mathcal{J}}} F_{\rm pdf}(\bar{J}) \log F_{\rm pdf}(\bar{J}) {\rm d}v_{\bar{\mathcal{J}}}(\bar{J}).
\end{align}
\end{define}

\begin{define}\label{def:info_gain}
For the $t$-th segment, let $F_{\rm pdf}^{t,k}(\bar{J})$ denote the posterior distribution updated using the first $k$ data points of the segment, and $F_{\rm pdf}^{t,k-1}(\bar{J})$ denote the posterior distribution updated using the first $k-1$ data points. The information gain at time $k$ within segment $t$ is defined as
\begin{align}\label{eq:info_gain}
I_t(k) &:= D_{\rm KL}(F_{\rm pdf}^{t,k} || F_{\rm pdf}^{t,k-1})\notag\\
 &= \int_{\bar{\mathcal{J}}} F_{\rm pdf}^{t,k}(\bar{J}) \log\left(\frac{F_{\rm pdf}^{t,k}(\bar{J})}{F_{\rm pdf}^{t,k-1}(\bar{J})}\right) {\rm d}v_{\bar{\mathcal{J}}}(\bar{J}).
\end{align}
\end{define}

According to the above definitions, the sampling stopping criterion can be described as:
\begin{align}
k \geq K_{\min} \quad \text{and} \quad \big(I_t(k) < \epsilon_I \ \text{or} \ k = K_{\max}\big), \label{eq:stop}
\end{align}
where $\epsilon_I > 0$ is a user-defined information gain threshold. This criterion ensures that segments are long enough to provide meaningful learning ($k \geq K_{\min}$) but allows early stopping when further data provides little new information ($I_t(k) < \epsilon_I$), or the segment length reaches its maximum ($k = K_{\max}$).

\begin{algorithm}[htbp]
\caption{TS-based ALC (adaptive segment length).}
\label{alg:TS}
\begin{algorithmic}[1]
\STATE {\textbf{Input}: prior distribution $F_{\rm pdf}^0(\bar{J})$; lower and upper segment bounds $K_{\min}, K_{\max}$; information gain threshold $\epsilon_I$;}\label{line:alg_input}
\STATE Randomly sample a function $\bar{J}_0$ with initial pdf $F_{\rm pdf}^0(\bar{J})$; \label{line:sample_barj0}
\STATE Select an initial control law by solving optimization problem ${\bm g}_0 = \arg\max\limits_{{\bm g}\in\mathcal{G}} \bar{J}_0({\bm g})$;\label{line:init_policy}
\STATE Apply the control law ${\bm g}_0$ to the system and begin collecting sample data for the first segment; \label{line:alg_apply_g0}
\STATE Initialize $R_0=1$;\label{line:initial}
\FOR{each iteration $t=1,2,\ldots$} \label{line:outer_for}
    \STATE $k \gets 0$; \label{line:alg_init_k}
    \STATE Initialize data buffer for current segment; \label{line:init_buffer}
    \STATE Observe initial state ${\bm x}_t(0)$; \label{line:observe_x0}
    \WHILE{$k < K_{\max}$} \label{line:while_k}
        \STATE Apply controller ${\bm g}_{t-1}$ to the system and observe data at time $k$; \label{line:apply_policy}
        \STATE Append observed data to current segment; \label{line:append_data}
        \IF{$k \geq K_{\min}$} \label{line:if_kmin}
            \STATE Compute information gain $I_t(k)$; \label{line:compute_ig}
            \STATE \textbf{If stopping criterion \eqref{eq:stop} is met:} \label{line:if_stop}
                \STATE \quad Stop current segment data collection; \label{line:stop_collect}
                \STATE \quad Calculate control cost $J_{t-1}({\bm g}_{t-1},{\bm x}_{t-1}(0))$ using data from current segment; \label{line:calc_cost}
                \STATE \quad Update the posterior distribution:\label{line:updis}
                \begin{align*}
                    R_t&=\frac{\bar{J}({\bm g}_{t-1};{\bm x}_{t-1}(0))}{\bar{J}^*({\bm g}_{t-1};{\bm x}_{t-1}(0))}R_{t-1};\\
                    F_{\rm pdf}^{t}(\bar{J})&=\frac{R_tF_{\rm pdf}^{0}(\bar{J})}{\int_{\bar{\mathcal{J}}}R_tF_{\rm pdf}^0({\rm d}v_{\bar{\mathcal{J}}}(\bar{J}))};
                \end{align*}
                \STATE \quad Randomly sample $\bar{J}_t$ according to posterior distribution $F_{\rm pdf}^{t}(\bar{J})$;\label{line:sample}
                \STATE \quad Select the controller ${\bm g}_t$ that maximizes $\bar{J}_t$;\label{line:cont}
                \STATE \quad Begin collecting data for next segment with ${\bm g}_t$; \label{line:begin_next_seg}
                \STATE \quad \textbf{break}; \label{line:alg_break}
        \ENDIF
        \STATE $k \gets k+1$; \label{line:inc_k}
    \ENDWHILE
    \IF{segment ended because $k = K_{\max}$} \label{line:if_max_k}
        \STATE Calculate control cost $J_{t-1}({\bm g}_{t-1},{\bm x}_{t-1}(0))$ using data from current segment; \label{line:calc_cost_max}
        \STATE Update posterior, sample $\bar{J}_t$, select ${\bm g}_t$ as above; \label{line:update_g_max}
        \STATE Begin next segment with ${\bm g}_t$; \label{line:begin_next_seg_max}
    \ENDIF
\ENDFOR
\end{algorithmic}
\end{algorithm}

Based on the distribution function $F_{\rm pdf}^{t}(\bar{J})$, the detailed procedure of the TS adaptive segment-length learning algorithm is illustrated in Algorithm \ref{alg:TS}. The algorithm first sets the prior distribution, samples the performance function, and selects the initial control law. Afterwards, the algorithm operates in a round-based loop. At the beginning of each round, it initializes the data buffer and observes the initial state. Then, data collection is performed stepwise in $k$, with the segment length limited by $K_{\max}$. At each step, the current control law is applied and observed results are stored in the buffer. Once the data length reaches $K_{\min}$, the algorithm calculates the current information gain $I_t(k)$ and checks whether the stopping criterion is met (i.e., when information gain falls below the threshold $\epsilon_I$ or the number of steps reaches $K_{\max}$). Upon meeting the stopping criterion, current segment data collection is terminated, the control cost of the previous round is calculated, and the posterior distribution $F_{\rm pdf}^t$ and weight $R_t$ are updated accordingly. Next, a new sample $\bar{J}_t$ is drawn from the updated posterior, and the new control law ${\bm g}_t$ is selected based on this sample to start the next round. If the round ends because the maximum step $K_{\max}$ is reached, the cost calculation, posterior update, and new control law selection are performed similarly before entering the next round. Overall, by adaptively adjusting the data segment length according to posterior uncertainty and continuously updating distributions and control policies in each round, this algorithm strikes a balance between exploration and exploitation, providing a systematic approach for end-to-end data-driven control without any structural assumptions on the model.


\subsubsection{Corollaries of Learning Convergence and Control Regret}

\begin{cor}\label{cor:conv}
	Suppose the controller sampling policy is implemented according to \eqref{eq:stop}. Then the predictive density $\bar{J}^T$ obtained in \eqref{eq:JTO} satisfies
	\begin{align}
	&\exists \epsilon_L,P_0,\delta_K, T_{\rm eff}:= \sum_{t=1}^T \mathbb{E}[K_t]/K_{\min}:\\
	&\mathbb{P}(\bar{J}^T({\bm g};{\bm x}_0)\in{\mathcal{B}}^{c}(\bar{J}^*,\delta_{\bar{J}}))\!<\!P_0\exp(-T_{\rm eff}\epsilon_L) \!+\! \delta_K, \text{a.s.}\notag
	\end{align}
	if the following conditions hold:
	\begin{align}
		&\lim\limits_{t\rightarrow\infty}\inf d(\bar{J}^t_\Omega,\bar{J}^*)>0,~\text{where}~\Omega=\mathcal{B}^{\rm c}(\bar{J}^*,\delta_{\bar{J}});\label{eqcor:dda0}\\
		&\sum\limits_{t=1}^T t^{-2}\mathbb{V}(T_d^t) < \infty;\\
		&\mathbb{E}[K_t] \geq K_{\min} > 0, \quad \mathbb{V}[K_t] < \infty, \quad \forall t.
	\end{align}
\end{cor}

	\begin{pf}
	For the considered $\sigma$-field $\Sigma_{\bar{{J}}}^{t-1}$ and Hellinger distance $D_{\rm H}(\cdot,\cdot)$, we have:
	\begin{align}
	&\mathbb{E}\left[\left.\sqrt{\frac{L_{t}(\Omega)}{L_{t-1}(\Omega)}}-1\right|\Sigma_{\bar{J}}^{t-1}\right]\label{eqcor:EcalH}\\
	=&\mathbb{E}\left[\left.\mathbb{E}_{{\bm x}_t(0)}\left[\sqrt{\frac{{\bar{J}^{t-1}}({\bm g}_{t};{\bm x}_t(0))}{\bar{J}^*({\bm g}_{t};{\bm x}_t(0))}}\right]-1\right|\Sigma_{\bar{J}}^{t-1}\right]\notag\\
	\leq& -\frac{1}{2}\mathbb{E}_{{\bm x}_t(0)}[D_{\rm H}^2({\bar{J}^{t-1}}(\cdot;{\bm x}_t(0)),\bar{J^*}(\cdot;{\bm x}_t(0)))],\notag
	\end{align}
	where the inequality follows from Jensen's inequality and the concavity of the square root function.
	
	Similarly, for KL divergence $D_{\rm KL}(\cdot,\cdot)$, we have
	\begin{align}
	&\mathbb{E}\left[\left.\log\frac{L_{t}(\Omega)}{L_{t-1}(\Omega)}\right|\Sigma_{\bar{J}}^{t-1}\right]\\
	=&\mathbb{E}\left[\left.\mathbb{E}_{{\bm x}_t(0)}\left[\log\frac{\bar{J}^{t-1}({\bm g}_{t};{\bm x}_t(0))}{\bar{J}^*({\bm g}_{t};{\bm x}_t(0))}\right]\right|\Sigma_{\bar{J}}^{t-1}\right]\notag\\	
	&\leq -\mathbb{E}_{{\bm x}_t(0)}[D_{\rm KL}(\bar{J}^{t-1}(\cdot;{\bm x}_t(0))||\bar{J^*}(\cdot;{\bm x}_t(0)))].\label{eqcor:EcalDKL}
	\end{align}
	where the inequality follows from Jensen's inequality and the convexity of the negative logarithm.

	By recalling the definition of $T_d(\cdot)$ in \eqref{eq:defTd} and accounting for the expectation over initial states, the following equation can be verified for both Hellinger distance and KL divergence:
	\begin{align}\label{eqcor:Emd}
	\mathbb{E}\left[\left.T_d\left(\frac{L_{t}(\Omega)}{L_{t-1}(\Omega)}\right)\right|\Sigma_{\bar{{J}}}^{t-1}\right]\leq -d(\bar{J}^{t-1}({\bm g}),\bar{J}^*({\bm g})) - \delta_{\rm init}^t,
	\end{align}
	where $\delta_{\rm init}^t$ accounts for the additional uncertainty due to the unknown initial state distribution, and satisfies $\delta_{\rm init}^t \geq 0$ with $\lim_{t\to\infty} \delta_{\rm init}^t = 0$ under the learning process.

	Then according to the definition of $T_d^t$ defined in \eqref{eq:defTdt}, equation \eqref{eqcor:Emd} can be equivalently rewritten as the following form for both Hellinger distance and KL divergence:
	\begin{align}
	\mathbb{E}[T_d^t|\Sigma_{\bar{J}}^{t-1}]+d(\bar{J}^{t-1}(\cdot),\bar{J}^*(\cdot))+\delta_{\rm init}^t\leq 0.
	\end{align}
	
	To construct a martingale, we define a predictable compensator process $A_T$ as follows:
	\begin{align}\label{eqcor:compensator}
	A_T = \sum_{t=1}^T \frac{K_t}{K_{\min}} \mathbb{E}\left[T_d^t+d(\bar{J}^{t-1}(\cdot),\bar{J}^*(\cdot))+\delta_{\rm init}^t \Big| \Sigma_{\bar{J}}^{t-1}\right],
	\end{align}
	where $\Sigma_{\bar{J}}^{0}$ is the trivial $\sigma$-field. Note that $A_T$ is predictable (i.e., $A_T$ is $\Sigma_{\bar{J}}^{T-1}$-measurable) and non-decreasing since the conditional expectation in \eqref{eqcor:Emd} is non-positive.
	
	Now, we redefine the martingale $M_T$ as:
	\begin{align}\label{eqcor:martingale_def}
	M_T=\sum\limits_{t=1}^T\left[\frac{K_t}{K_{\min}}\left(T_d^t+d(\bar{J}^{t-1}(\cdot),\bar{J}^*(\cdot))+\delta_{\rm init}^t\right)\right] - A_T.
	\end{align}
	
	We claim that:
	\begin{align}
	&\mathbb{E}\left[|M_T|\right]<\infty,\label{eqcor:mat_1}\\
	&\mathbb{E}\left[M_{T+1}|M_1,\ldots,M_T\right]=M_T.\label{eqcor:mat_2}
	\end{align}

	For \eqref{eqcor:mat_1}, by the definition of $K_t$, we know that $K_t \geq K_{\min} > 0$, and as established earlier, $K_t$ is bounded and $\mathbb{E}[K_t]<\infty$. Furthermore, $T_d^t$ and $d(\bar{J}^{t-1}(\cdot),\bar{J}^*(\cdot))$ correspond to the Hellinger distance or KL divergence, and are therefore both nonnegative and bounded. The term $\delta_{\rm init}^t$ is also a nonnegative quantity that converges to zero. The compensator $A_T$ is also bounded since it is a sum of bounded conditional expectations. As a result, $|M_T|$ is bounded, so its expectation is bounded, i.e., $\mathbb{E}\left[|M_T|\right]<\infty$.
	
	For \eqref{eqcor:mat_2}, by the definition of $M_T$ in \eqref{eqcor:martingale_def}, we have:
	\begin{align}
	&M_{T+1} - M_T \\
	=& \frac{K_{T+1}}{K_{\min}}\left(T_d^{T+1}+d(\bar{J}^{T}(\cdot),\bar{J}^*(\cdot))+\delta_{\rm init}^{T+1}\right) \notag\\
	&\quad - \left(A_{T+1} - A_T\right)\notag\\
	=& \frac{K_{T+1}}{K_{\min}}\left(T_d^{T+1}+d(\bar{J}^{T}(\cdot),\bar{J}^*(\cdot))+\delta_{\rm init}^{T+1}\right) \notag\\
	&\quad - \frac{K_{T+1}}{K_{\min}} \mathbb{E}\left[T_d^{T+1}+d(\bar{J}^{T}(\cdot),\bar{J}^*(\cdot))+\delta_{\rm init}^{T+1} \Big| \Sigma_{\bar{J}}^{T}\right]\notag.
	\end{align}
	Taking conditional expectation with respect to $\Sigma_{\bar{J}}^{T}$ (which contains $M_1,\ldots,M_T$), we obtain:
	\begin{align}
	&\mathbb{E}[M_{T+1} - M_T | \Sigma_{\bar{J}}^{T}]=0.
	\end{align}
	Therefore, $M_T$ satisfies the martingale property, i.e., $\mathbb{E}\left[M_{T+1}|M_1,\ldots,M_T\right]=M_T$.

	Thus, the stochastic process $M_T$ is a martingale. Therefore, according to the martingale convergence theorem \cite{yau2024principles}, we have
	\begin{align}\label{eqcor:limMTT}
	\lim\limits_{T\rightarrow\infty}\frac{M_T}{T_{\rm eff}}=0,~\text{a.s.},
	\end{align}
	where $T_{\rm eff} = \sum_{t=1}^T \mathbb{E}[K_t]/K_{\min}$.
	
	From the definition of $M_T$ in \eqref{eqcor:martingale_def}, we have:
	\begin{align}
	\frac{M_T}{T_{\rm eff}} = \frac{1}{T_{\rm eff}}\sum\limits_{t=1}^T\frac{K_t}{K_{\min}}\left[T_d^t+d(\bar{J}^{t-1}(\cdot),\bar{J}^*(\cdot))+\delta_{\rm init}^t\right] - \frac{A_T}{T_{\rm eff}}.
	\end{align}
	Since $M_T/T_{\rm eff} \to 0$ a.s. and the compensator $A_T$ is predictable and non-decreasing with bounded increments (due to equation \eqref{eqcor:Emd}), we have that $A_T/T_{\rm eff}$ converges to a finite limit. Therefore, equation \eqref{eqcor:limMTT} can be equivalently rewritten as
	\begin{align}\label{eqcor:Td0}
	&\lim\limits_{T\rightarrow\infty}\frac{1}{T_{\rm eff}}\sum\limits_{t=1}^T\frac{K_t}{K_{\min}}\left[T_d\left(\frac{L_{t}(\Omega)}{L_{t-1}(\Omega)}\right)\!+\!d(\bar{J}^t(\cdot),\bar{J}^*(\cdot))\!+\!\delta_{\rm init}^t\right]\notag\\
	=&\lim\limits_{T\rightarrow\infty}\frac{A_T}{T_{\rm eff}},~\text{a.s.}
	\end{align}
	Since $A_T$ is the sum of non-positive conditional expectations (from \eqref{eqcor:Emd}), we have $A_T \leq 0$ and thus $\lim_{T\to\infty} A_T/T_{\rm eff} \leq 0$. Combined with condition \eqref{eqcor:dda0}, this implies that the limit in \eqref{eqcor:Td0} is actually zero, i.e.,
	\begin{align}\label{eqcor:Td0_actual}
	&\lim\limits_{T\rightarrow\infty}\frac{1}{T_{\rm eff}}\sum\limits_{t=1}^T\frac{K_t}{K_{\min}}\left[T_d\left(\frac{L_{t}(\Omega)}{L_{t-1}(\Omega)}\right)\!+\!d({\bar{J}^{t-1}}(\cdot),\bar{J}^*(\cdot))\!+\!\delta_{\rm init}^t\right]\notag\\
	&=0,~\text{a.s.}
	\end{align}
	
	By combining equations \eqref{eqcor:dda0} and \eqref{eqcor:Td0_actual}, we have
	\begin{align}\label{eqcor:supT1T}
	\sup\limits_{T}\frac{1}{T_{\rm eff}}\sum\limits_{t=1}^T\frac{K_t}{K_{\min}}T_d\left(\frac{L_{t}(\Omega)}{L_{t-1}(\Omega)}\right)<0.
	\end{align}
	According to equation \eqref{eqcor:supT1T}, the following claim can be proved:
	\begin{align}\label{eqcor:claiminTHM1}
	\exists\epsilon_L(\Omega):~L_T(\Omega)<L_0(\Omega)\exp(-T_{\rm eff}\epsilon_L) + \delta_K, \text{a.s.},
	\end{align}
	where $\delta_K$ accounts for the variance in segment lengths and satisfies:
	\begin{align}
	\delta_K = \mathcal{O}\left(\frac{\mathbb{V}[K_t]}{\mathbb{E}[K_t]^2}\right).
	\end{align}
	
	Then, we have
	\begin{align}
	\lim\limits_{t\rightarrow\infty}\mathbb{P}(\bar{J}\in\Omega)\leq \delta_K,~\text{a.s.}
	\end{align}
	As a result, by defining 
	\begin{align}
	P_0=\frac{L_0(\Omega)}{\max\limits_{t}\int_{\bar{\mathcal{J}}}R_tF_{\rm pdf}^0({\rm d}{v_{\bar{\mathcal{J}}}(\bar{J})})},
	\end{align}
	we obtain 
	\begin{align}
	\mathbb{P}(\bar{J}\in\Omega)<P_0\exp(-T_{\rm eff}\epsilon_L) + \delta_K,~\text{a.s.},
	\end{align}
	which completes the proof of Corollary \ref{cor:conv}. \hfill\QEDopen
	\end{pf}

\begin{cor}\label{cor:regret}
		Let Assumptions \ref{assum:Xbar}-\ref{assum:Gstar} hold. With the proposed parameterization approach in \eqref{eq:Gsumm} and the implementation of the TS-based learning in Algorithm \ref{alg:TS}, the regret defined in \eqref{eq:regretdefine} satisfies the following upper bound:
		\begin{align}\label{eq:regretbound}
		&R_e({\bm g}_t)\\
		\leq&\bar{J}_{\rm max}v_{\bar{\mathcal{J}}}(\mathcal{B}(\bar{J}^*,\delta_{\bar{J}}))+\frac{\epsilon_J L_J\sqrt{mM_g^2-\sum_{i=1}^m\|{ g}_i^*\|_{\mathcal{L}_2}^2}}{\sqrt{2^n}J_{\rm min}^2}\notag\\
		&+\left[2\bar{M}_{\bar{\mathcal{J}}}-\bar{J}_{\rm max}v_{\bar{\mathcal{J}}}(\mathcal{B}(\bar{J}^*,\delta_{\bar{J}}))\right]P_0\exp(-t_{\rm eff}\epsilon_L)\notag\\
		&+\delta_{\rm init} \cdot L_J \cdot \mathbb{V}[{\bm x}(0)] + \delta_K \cdot \frac{\mathbb{V}[K_t]}{\mathbb{E}[K_t]^2},~\text{a.s.},\notag
		\end{align}
		where $t_{\rm eff} = \sum_{i=1}^t \mathbb{E}[K_i]/K_{\min}$ is the effective segment index, $\delta_{\rm init}$ is a constant related to the initial state distribution uncertainty, $\mathbb{V}[{\bm x}(0)]$ is the variance of the initial state distribution, and $\epsilon_J>0$ is a constant.
\end{cor}

\begin{pf}
	The proof of Corollary \ref{cor:regret} can be conducted using the same approach as in Theorem \ref{thm:regret}. The main difference compared to Theorem \ref{thm:regret} is that the results from Theorem \ref{thm:conv} used in \eqref{eq:term11} and \eqref{eq:term12} should be replaced by the corresponding results from Corollary \ref{cor:conv}.
\end{pf}

		\subsubsection{Extension on Nonstationary Reward Distribution}
		Although this work mainly focuses on stationary processes, the proposed method can be extended to address nonstationary processes, such as slowly time-varying system dynamics or changes in the cost function due to control objective variation. The proposed algorithm can adapt to nonstationary processes by continuously updating the posterior distribution with newly acquired data. In scenarios where the environment changes rapidly, sliding window mechanisms \cite{trovo2020sliding} or discounting mechanisms \cite{raj2017taming} can reduce the influence of historical data on the posterior distribution, thus enhancing learning efficiency. 
	For instance, one possible approach is to introduce a discounting factor $\lambda_{\rm f}\in(0,1)$ to control the impact of historical data on the posterior distribution:
	\begin{align}
		R_t = \lambda_{\rm f} \cdot \frac{\bar{J}({\bm g}_{t-1})}{\bar{J}^*({\bm g}_{t-1})}R_{t-1} + (1-\lambda_{\rm f}) \cdot \frac{\bar{J}({\bm g}_{t-1})}{\bar{J}^*({\bm g}_{t-1})}.
	\end{align}
	However, in this case, the convergence of the proposed learning algorithm will be affected by a combination of factors including the discounting factor, learning efficiency, and the degree of environmental variation, making the analysis of specific theoretical results challenging.

	\subsection{Closed-Loop Stability Analysis}
		The method proposed in this paper is an approach to learning controllers for unknown dynamical systems through online active exploration. The closed-loop stability of the system under the proposed learning strategy can be discussed from two perspectives: (1) during the learning process itself, and (2) after a period of exploration.
	
		During the learning process, it is possible that some unstable control laws may be used, since the proposed method utilizes very little prior information. To address such situations, it is recommended to define a safe control policy, ${\bm g}_{\rm safe}$, which can be determined by other controller design techniques or historical test data. During online learning, if the system state exceeds a predefined safe region $\mathcal{X}_{\rm safe}$, the controller ${\bm g}_{\rm safe}$ should be applied. Controllers that cause instability in such cases will be assigned a sufficiently large cost (or equivalently, a sufficiently small reward). In this way, potential safety issues during the learning process can be avoided.

	After a period of exploration, the closed-loop stability of the system can be guaranteed with high probability, which is analyzed in the following results.

\begin{assumption}\label{assum:stable}
	It is assumed that the optimal control law ${\bm g}^*$ renders the closed-loop system mean-square bounded; namely, there exists a constant $M^*>0$ such that $\sup_{k\geq 0}\mathbb{E}[\|{\bm x}^*(k)\|^2]\leq M^*$, where ${\bm x}^*(k)$ denotes the state trajectory under the optimal control law ${\bm g}^*$.
\end{assumption}

\begin{thm}\label{thm:mean_square_stability}
	Let Assumptions \ref{assum:Xbar}-\ref{assum:Gstar} and \ref{assum:stable} hold. Consider the system \eqref{eq:sys} with the cost function $J$ defined in \eqref{eq:costJ}. The closed-loop system under the learned control law ${\bm g}_t$ is mean-square bounded with probability at least $1-P_0\exp(-t\epsilon_L)$, i.e., there exists a constant $M>0$ such that
	\begin{align}\label{eq:mean_square_bound}
		\sup_{0\leq k\leq K}\mathbb{E}[\|{\bm x}_t(k)\|^2]\leq M,
	\end{align}
	where ${\bm x}_t(k)$ denotes the state trajectory in the $t$-th segment under the control law ${\bm g}_t$.
\end{thm}

\begin{pf}
	The proof consists of three steps: (1) establishing the relationship between the cost function and state norms, (2) deriving an upper bound on the expected cost using regret analysis, and (3) proving mean-square boundedness of the state.
	\newline
	\noindent\textbf{Step 1: Cost function and state norm relationship.}~
	According to the cost function definition in \eqref{eq:sys}, for the $t$-th segment with length $K$, we have
	\begin{align}\label{eq:cost_state}
		J_t({\bm g}_t)=\sum\limits_{k=0}^{K}({\bm x}_t(k)^{\rm T}{ Q}{\bm x}_t(k)+{\bm u}_t(k)^{\rm T}{ R}{\bm u}_t(k)),
	\end{align}
where ${\bm x}_t(k)$ and ${\bm u}_t(k)$ are the state and control input in the $t$-th segment, respectively.

	Since ${ Q}$ is positive definite, there exists $\lambda_{\min}({ Q})>0$ such that ${\bm x}^{\rm T}{ Q}{\bm x}\geq \lambda_{\min}({ Q})\|{\bm x}\|^2$ for all ${\bm x}\in\mathbb{R}^n$. Therefore, we have
	\begin{align}\label{eq:cost_lower_bound}
		J_t({\bm g}_t)\geq \lambda_{\min}({ Q})\sum\limits_{k=0}^{K}\|{\bm x}_t(k)\|^2.
	\end{align}
	Similarly, for the optimal control law ${\bm g}^*$, we have
	\begin{align}\label{eq:cost_optimal}
		J^*({\bm g}^*)=\sum\limits_{k=0}^{K}({\bm x}^*(k)^{\rm T}{ Q}{\bm x}^*(k)+{\bm u}^*(k)^{\rm T}{ R}{\bm u}^*(k)),
	\end{align}
	and
	\begin{align}\label{eq:cost_optimal_lower}
		J^*({\bm g}^*)\geq \lambda_{\min}({ Q})\sum\limits_{k=0}^{K}\|{\bm x}^*(k)\|^2.
	\end{align}
	\noindent\textbf{Step 2: Upper bound on expected cost.}~
	According to the regret definition in \eqref{eq:regretdefine} as $R_e({\bm g}_t)=\left|\mathbb{E}\left[\bar{J}^*({\bm g}^*)\right]-\mathbb{E}\left[\bar{J}_t({\bm g}_t)\right]\right|$ and the relationship between the reward function $\bar{J}$ and the cost function $J$ in \eqref{eq:prop}, we have
	\begin{align}\label{eq:regret_cost}
		R_e({\bm g}_t)=\left|\mathbb{E}\left[\bar{J}^*({\bm g}^*)\right]-\mathbb{E}\left[\bar{J}_t({\bm g}_t)\right]\right|.
	\end{align}
	From the relationship between cost functions $J^*, J_t$ and reward functions $\bar{J}^*, \bar{J}_t$, we know that $\bar{J}^*({\bm g}^*)\propto\frac{1}{J^*({\bm g}^*)}$ and $\bar{J}_t({\bm g}_t)\propto\frac{1}{J_t({\bm g}_t)}$. Let the proportional coefficient be $\epsilon_J>0$ as in Theorem \ref{thm:regret}, then we have $\bar{J}^*({\bm g}^*)=\frac{\epsilon_J}{J^*({\bm g}^*)}$ and $\bar{J}_t({\bm g}_t)=\frac{\epsilon_J}{J_t({\bm g}_t)}$.
	\newline
	According to Theorem \ref{thm:regret}, the regret $R_e({\bm g}_t)$ is bounded as
	\begin{align}\label{eq:regret_bound_used}
		R_e({\bm g}_t)\leq&\bar{J}_{\rm max}v_{\bar{\mathcal{J}}}(\mathcal{B}(\bar{J}^*,\delta_{\bar{J}}))+\frac{\epsilon_J L_J\sqrt{mM_g^2-\sum_{i=1}^m\|{ g}_i^*\|_{\mathcal{L}_2}^2}}{\sqrt{2^n}J_{\rm min}^2}\notag\\
		&+\left[2\bar{M}_{\bar{\mathcal{J}}}-\bar{J}_{\rm max}v_{\bar{\mathcal{J}}}(\mathcal{B}(\bar{J}^*,\delta_{\bar{J}}))\right]P_0\exp(-t\epsilon_L),\\
	    &=:C_R,\notag
	\end{align}
	which can be rewritten as
	\begin{align}\label{eq:regret_ineq}
		\left|\mathbb{E}\left[\frac{\epsilon_J}{J^*({\bm g}^*)}\right]-\mathbb{E}\left[\frac{\epsilon_J}{J_t({\bm g}_t)}\right]\right|\leq C_R.
	\end{align}
	By rearranging \eqref{eq:regret_ineq}, we obtain
	\begin{align}
		&\mathbb{E}\left[\frac{1}{J_t({\bm g}_t)}\right]\geq \mathbb{E}\left[\frac{1}{J^*({\bm g}^*)}\right]-\frac{C_R}{\epsilon_J},\label{eq:cost_upper_bound1}\\
		&\mathbb{E}\left[\frac{1}{J_t({\bm g}_t)}\right]\leq \mathbb{E}\left[\frac{1}{J^*({\bm g}^*)}\right]+\frac{C_R}{\epsilon_J}.\label{eq:cost_upper_bound2}
	\end{align}
	Since $J({\bm g})\geq J_{\rm min}>0$ for all ${\bm g}\in\mathcal{G}$ (from Assumption \ref{assum:LipofJ}), and the regret in terms of $\bar{J}$ is bounded, we can show that there exists a constant $C_J>0$ such that
	\begin{align}\label{eq:cost_upper_final}
		\mathbb{E}[J_t({\bm g}_t)]\leq \mathbb{E}[J^*({\bm g}^*)]+\frac{C_R \bar{J}_{\rm max}^2}{\epsilon_J J_{\rm min}^2}.
	\end{align}
	The constant $C_J$ can be expressed as $C_J=\frac{C_R \bar{J}_{\rm max}^2}{\epsilon_J J_{\rm min}^2}$, where $\bar{J}_{\rm max}$ is the upper bound of $\bar{J}$ (from property 5 in the definition of $\bar{\mathcal{J}}$).
	\newline
	\noindent\textbf{Step 3: Mean-square boundedness.}~
	Taking expectation on both sides of \eqref{eq:cost_lower_bound}, we obtain
	\begin{align}\label{eq:expect_cost_lower}
		\mathbb{E}[J_t({\bm g}_t)]\geq \lambda_{\min}({ Q})\sum\limits_{k=0}^{K}\mathbb{E}[\|{\bm x}_t(k)\|^2].
	\end{align}
	Combining \eqref{eq:cost_upper_final} and \eqref{eq:expect_cost_lower}, we have
	\begin{align}\label{eq:state_sum_bound}
		\lambda_{\min}({ Q})\sum\limits_{k=0}^{K}\mathbb{E}[\|{\bm x}_t(k)\|^2]\leq \mathbb{E}[J^*({\bm g}^*)]+C_J.
	\end{align}
	Then according to Assumption \ref{assum:stable}, we have
	\begin{align}\label{eq:optimal_cost_bound}
		\mathbb{E}[J^*({\bm g}^*)]\!=\!\mathbb{E}\left[\sum\limits_{k=0}^{K}{\bm x}^*(k)^{\rm T}{ Q}{\bm x}^*(k)\!+\!{\bm u}^*(k)^{\rm T}{ R}{\bm u}^*(k)\right]\!\leq\! C_{J^*},
	\end{align}
	where $C_{J^*}:=(K+1)\left[\lambda_{\rm max}(Q)M^*+\lambda_{\rm max}(R)\max\limits_{x}\|{\bm g}^*({\bm x})\|^2 \right]$ is a constant.
	\newline
	Substituting \eqref{eq:state_sum_bound} into \eqref{eq:expect_cost_lower}, we obtain
	\begin{align}\label{eq:state_sum_final}
		\sum\limits_{k=0}^{K}\mathbb{E}[\|{\bm x}_t(k)\|^2]\leq \frac{C_{J^*}+C_J}{\lambda_{\min}({ Q})}.
	\end{align}
	Since the right-hand side of \eqref{eq:state_sum_final} is independent of $k$, and the sum is over $K+1$ terms, we have
	\begin{align}\label{eq:state_mean_bound}
		\frac{1}{K+1}\sum\limits_{k=0}^{K}\mathbb{E}[\|{\bm x}_t(k)\|^2]\leq \frac{C_{J^*}+C_J}{\lambda_{\min}({ Q})(K+1)},
	\end{align}
	which leads to the mean-square boundedness of the state as $\sup\limits_{0\leq k\leq K}\mathbb{E}[\|{\bm x}_t(k)\|^2]\leq \frac{C_{J^*}+C_J}{\lambda_{\min}({ Q})}$. \hfill\QEDopen
\end{pf}

\subsection{Computational Complexity Analysis}
The computational complexity of the proposed method can be examined from two main aspects: the computation of the feedback control ${\bm u}(k)={\bm g}_k({\bm x}(k))$ and the update of the control law ${\bm g}_k(\cdot)$. 

The complexity associated with computing the feedback control ${\bm u}(k)={\bm g}_k({\bm x}(k))$ is determined by the complexity of the initial control law $\check{\bm g}$. The complexities of evaluating the individual initial control laws $\check{g}_1, \ldots, \check{g}_m$ are denoted as $\mathcal{O}(C_{\check{g}_1}), \ldots, \mathcal{O}(C_{\check{g}_m})$ respectively. Then, according to the proposed decomposition scheme \eqref{eq:calGi}, the complexity of computing each basis function $g^{(i)}_{{\bm w},\tilde{\bm x}}$ does not exceed $\mathcal{O}(C_{\check{g}_i})$. For high-dimensional systems, since the control input for each channel typically depends on only a subset of the states, let $d_i$ denote the number of states directly influencing the computation of input ${\bm u}_i$. Consequently, the computational complexity of calculating the feedback control law ${\bm u}(k)={\bm g}_k({\bm x}(k))$ is $\sum\limits_{i=1}^{m}\mathcal{O}(2^{d_i}C_{\check{g}_i})$, as the function space $\mathcal{G}_i$ comprises $2^{d_i}$ basis functions. 

With respect to updating the control law ${\bm g}_k(\cdot)$, the computational complexity depends on the parameterization of the reward function $\bar{J}^t(\cdot)$. Although the update of the posterior probability density $F_{\rm pdf}^{t}(\bar{J})=\frac{R_tF_{\rm pdf}^{0}(\bar{J})}{\int_{\bar{\mathcal{J}}}R_tF_{\rm pdf}^0({\rm d}v_{\mathcal{J}}(\bar{J}))}$ involves a potentially intricate integral, this computation can be avoided by leveraging the normalization property of probability density functions. In this work, when the reward function is modeled using Gaussian processes and kernel decomposition techniques \cite{hernandez2014predictive}, the complexity of updating the reward function is $\mathcal{O}(M_{\rm kernel}^3 + T M_{\rm kernel})$ \cite{vakili2021scalable}, where $T$ denotes the number of segments and $M_{\rm kernel}$ is the number of eigenfunctions employed in the kernel decomposition of the Gaussian process.

\section{Implementation and Simulation}\label{sec:NumericalExperiments}
To illustrate the effectiveness of the proposed method, consider a nonlinear system with the following dynamics:
\begin{align*}
\dot{x}_1 &= x_2+v_1, \\
\dot{x}_2 &= x_3+v_2, \\
\dot{x}_3 &= \sin(x_1) -2x_2^2+2x_1x_2 + u + v_3,
\end{align*}
where ${\bm x} = [x_1, x_2, x_3]^\top$ represents the system state, $u$ denotes the control input, and $v_1,v_2,v_3$ are Gaussian noises with a standard deviation of $\sqrt{2}\times10^{-2}$. The initial state of the system is $x(0)=[0,0,0.3]^\top$. The Euler method is used to discretize the system dynamics, with a sampling period of $T_s=0.05$. The cost function for the system is designed as $J=\sum\limits_{k=1}^{K} \left( x(k)^\top Q x(k) + u(k)^\top R u(k) \right)$, where $Q$ is an identity matrix, and $R=0.1$. The initial controller is an unstable controller $u=\bar{\bm g}({\bm x}(k))=x_1+x_1^2+x_2^2+3x_1x_2$. In the designed numerical experiment, the length of each segment is set to $K=50$, and the parameter $\Gamma$ is chosen to be $5$. 

\begin{figure}[H]
	\centering
	\subfloat[Basis Function $g_1({\bm x})$]{\includegraphics[width=0.45\linewidth]{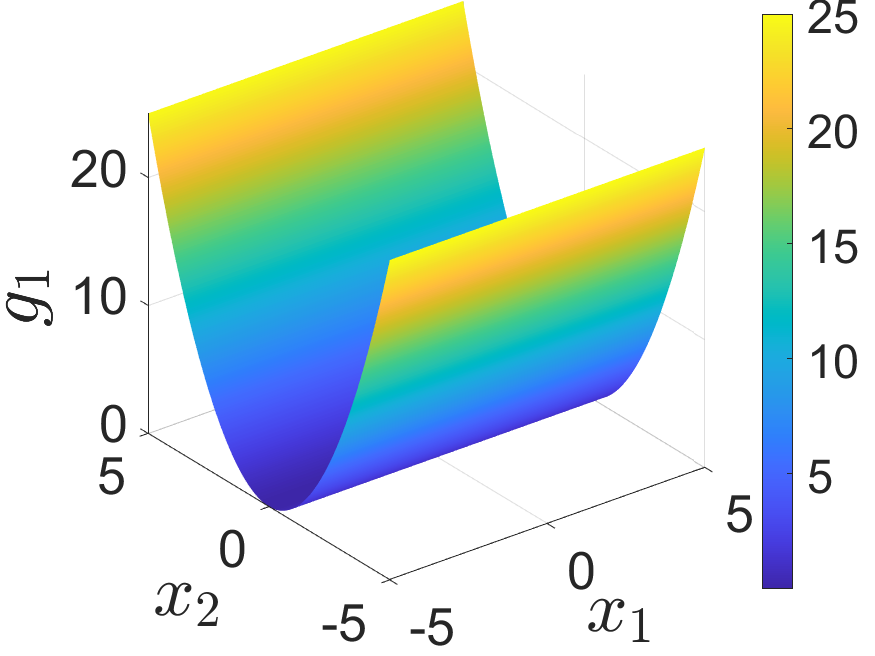}}
	\hfill
	\subfloat[Basis Function $g_2({\bm x})$]{\includegraphics[width=0.45\linewidth]{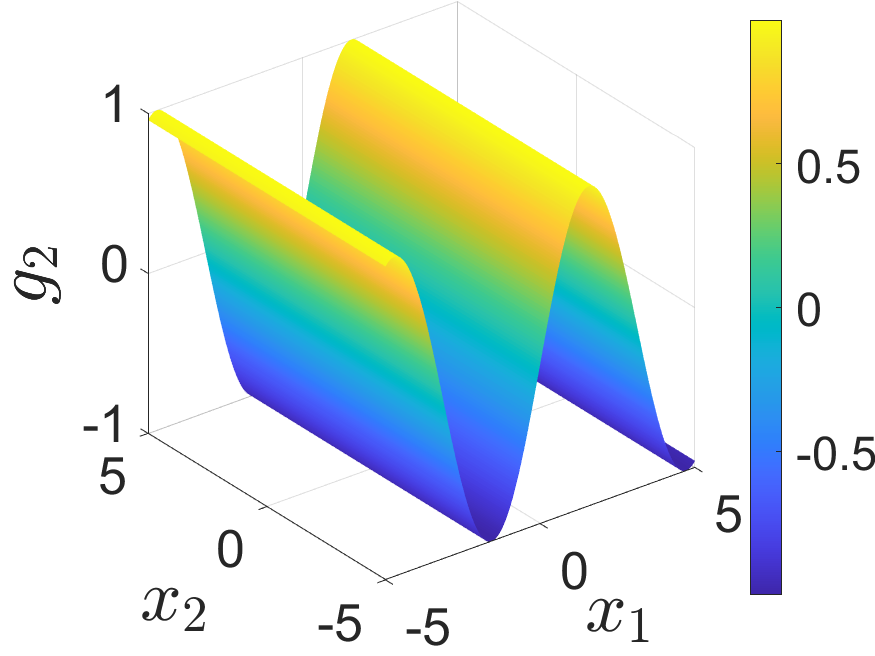}}
	\\
	\subfloat[Basis Function $g_3({\bm x})$]{\includegraphics[width=0.45\linewidth]{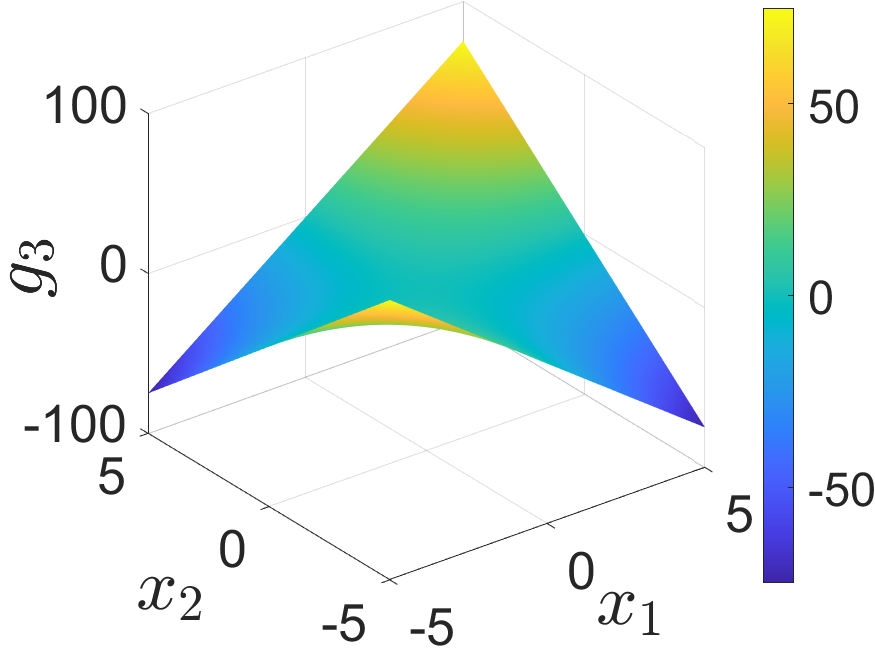}}
	\hfill
	\subfloat[Basis Function $g_4({\bm x})$]{\includegraphics[width=0.45\linewidth]{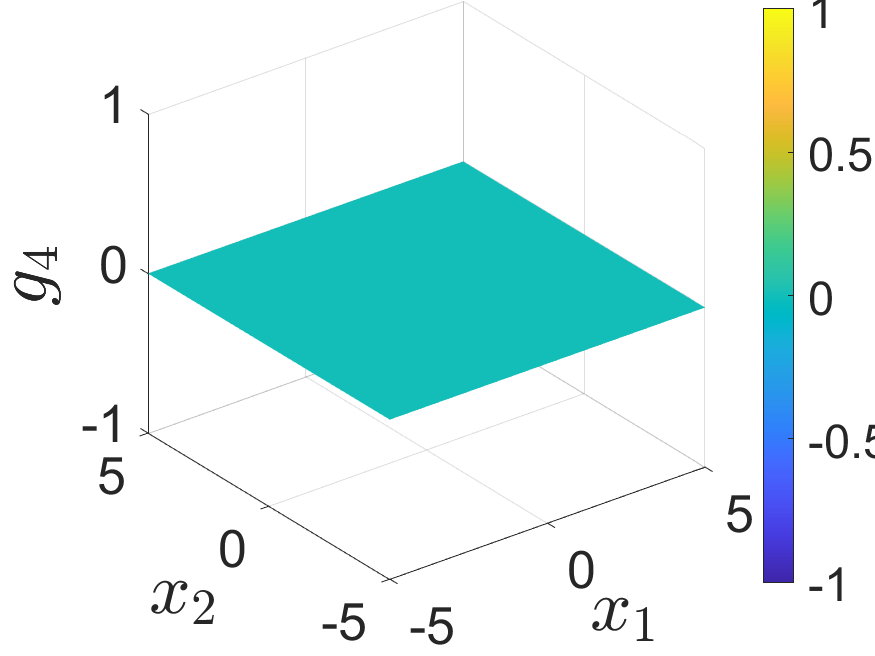}}
	\caption{Basis functions of the function space $\mathcal{G}$, which is obtained from the parameterization approach \eqref{eq:calGi}.}
	\label{fig:base_functions}
\end{figure}

According to \eqref{eq:calGi}, four basis functions are constructed, which are shown in Fig.~\ref{fig:base_functions}. In this example, although $n=3$, since the initial control law $\check{\bm g}(x)$ only depends on states $x_1$ and $x_2$, the number of actually effective basis functions is $2^2=4$.
The fourth basis function is a zero function because it represents the constant component of the function $\bar{\bm g}$, i.e., $g_4({\bm x})=\bar{\bm g}({\bm 0})=0$. Using the obtained basis functions and the corresponding constructed RKHS, we can search for the control law ${\bm g}_t$ in the candidate space $\mathcal{G}$ that maximizes the learned cost function $\bar{J}^t({\bm g})$. 

Two types of tests are conducted to verify the effectiveness of Theorems \ref{thm:conv} and \ref{thm:regret}, and their results are combined for analysis. To quantitatively evaluate the accuracy of the learned reward function $\bar{J}^t({\bm g})$, we first construct a reliable ground-truth benchmark by estimating the true reward function $\bar{J}^*({\bm g})$. Specifically, we eliminate system noise (set to zero) and uniformly sampled $8000$ candidate controllers by uniformly selecting control laws at fixed intervals in the coefficient space of $\mathcal{G}$, thereby generating the set $G_{\rm s}$. For each controller, the corresponding cost $J=\sum_{k=1}^{K} ( x(k)^\top Q x(k) + u(k)^\top R u(k) )$ is computed, and these results are used to construct the ground-truth reward function $\bar{J}^*({\bm g})$. Since statistical metrics such as the KL divergence $D_{\rm KL}(\bar{J}^t||\bar{J}^*)$ and the Hellinger distance $D_{\rm H}(\bar{J}^t||\bar{J}^*)$ are generally intractable to compute analytically in the continuous function space, we use these samples to numerically approximate the distances. By evaluating both $\bar{J}^t$ and $\bar{J}^*$ at the sampled points, we can effectively and efficiently evaluate the learning performance of the proposed approach.

The second type of test evaluates the proposed active learning method. In this test, 500 independent trials are conducted, with each trial consisting of 200 segments. At the end of the $t$-th segment in each trial, the data collected during the current segment is used to update the control law for the $(t+1)$-th segment according to Algorithm \ref{alg:TS}.

\begin{figure}[htbp]
	\centering
	\includegraphics[width=0.8\linewidth]{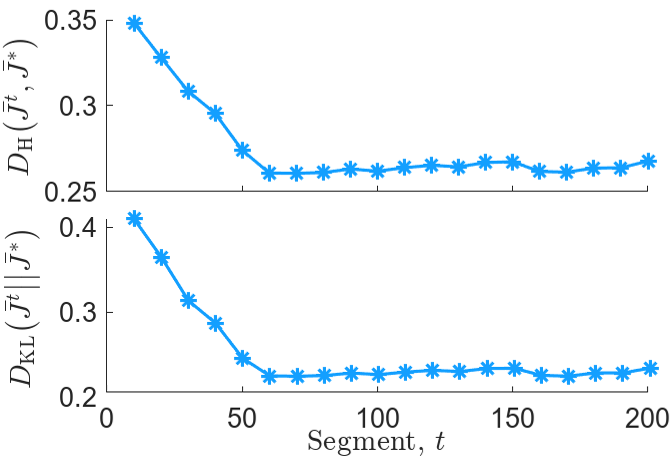}
\caption{Hellinger distance (top subplot) and KL divergence (bottom subplot) between the learned reward function $\bar{J}^t$ and the ground truth $\bar{J}^*$.}
	\label{fig:distance}
\end{figure}

\begin{figure}[htbp]
	\centering
	\includegraphics[width=0.8\linewidth]{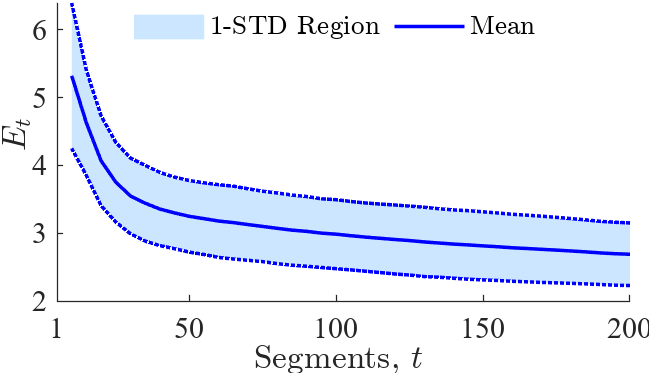}
\caption{Convergence of the learned reward function $\bar{J}^t$ as a function of time. The blue line represents the average value, while the shaded area with dashed lines represents the standard deviation deviating from the average value.}
	\label{fig:convergence}
\end{figure}

To verify the convergence of the learned reward function $\bar{J}^t$ as stated in Theorem \ref{thm:conv}, the comparison results are shown in Fig.~\ref{fig:distance} and Fig.~\ref{fig:convergence}. Specifically, Fig.~\ref{fig:distance} illustrates the KL divergence and Hellinger distance between the learned reward function $\bar{J}^t$ and the true reward function $\bar{J}^*$ in a single independent test. Fig.~\ref{fig:convergence} shows the average and standard deviation of the learning error $E_t$ over 500 independent trials as a function of time, where the learning error $E_t$ is defined as 
\begin{align}\label{eq:Er}
E_t:=\sum\limits_{g \in G_{\rm s}} \left(\bar{J}^t(g)-\bar{J}^*(g)\right)^2.
\end{align}

Figs.~\ref{fig:distance} and \ref{fig:convergence} show that the learned reward function $\bar{J}^t$ converges exponentially to the neighborhood of the true reward function $\bar{J}^*$, including the KL divergence, Hellinger distance, and the learning error defined in \eqref{eq:Er}. 
The results in Fig.~\ref{fig:distance} are consistent with Theorem \ref{thm:conv}, indicating that the learned reward function $\bar{J}^t(\cdot)$ enters a neighborhood $\mathcal{B}(\bar{J}^*(\cdot),\delta_{\bar{J}})$ of the true reward function $\bar{J}^*(\cdot)$. This is because the proposed method adopts an active sampling approach, which constantly explores other possibilities. The suboptimality can be characterized by the constant $\delta_{\bar{J}}$ in Theorem \ref{thm:conv}.
 These results validate the correctness of Theorem \ref{thm:conv}. 

\begin{figure}[htbp]
	\centering
	\includegraphics[width=0.8\linewidth]{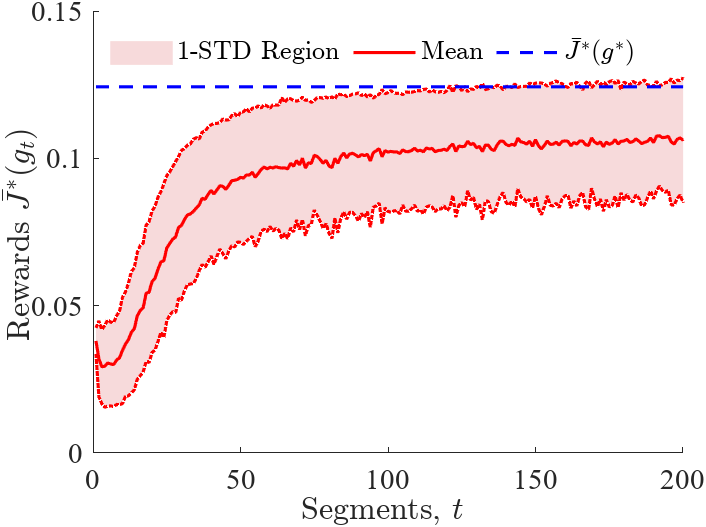}
\caption{Rewards during the proposed ALC processes. The red solid line represents the average of 500 independent trials, the light red area indicates the range within one standard deviation from the mean, and the blue dashed line represents the reward of the optimal control law under noise-free conditions.}
	\label{fig:regret}
\end{figure}

To verify the control regret in Theorem \ref{thm:regret}, the rewards corresponding to the control laws obtained from the TS algorithm in each independent trial are recorded. The mean and 1-standard deviation range (1-STD Region) of 500 independent trials are shown in Fig.~\ref{fig:regret}. For comparison, Fig.~\ref{fig:regret} also displays the reward corresponding to the optimal control law under noise-free conditions. It can be observed that the expected control reward converges exponentially to around 0.106, while the reward of the optimal control law is 0.1243. Fig.~\ref{fig:regret} validates the correctness of Theorem \ref{thm:regret}. Specifically, Fig.~\ref{fig:regret} shows that in 500 independent trials, the control reward increases exponentially during the first 50 segments (corresponding to the third term in equation \eqref{eq:regretbound}), but there remains a certain gap from the optimal reward (corresponding to the first and second terms in equation \eqref{eq:regretbound}).

By observing Figs.~\ref{fig:convergence} and \ref{fig:regret}, it can be seen that both the reward function $\bar{J}^t$ and the control reward exhibit similar convergence patterns. During the first 50 segments, the reward function and control reward converge rapidly to the vicinity of the true value at an exponential rate, while the convergence rate slows down in subsequent periods. This pattern is due to the balance between exploration and exploitation in the TS method. After the initial rapid exploration, the learned reward function $\bar{J}^t$ becomes closer to the true value $\bar{J}^*$, especially in regions with higher rewards (as these regions are sampled with higher probabilities and thus have more sampling points). This reduces the probability of sampling other regions with lower expected rewards. Therefore, overall, in the proposed active learning method, when the reward function $\bar{J}^t$ converges near the true value $\bar{J}^*$, the control law is close to the optimal control law ${\bm g}^*$ most of the time, but occasionally selects other control laws for exploration. Exploration in regions with lower rewards leads to the slow convergence of the reward function $\bar{J}^t$ and also results in a certain difference between the expected control reward and the optimal reward. Similar convergence patterns are also commonly observed in other TS-based methods, such as \cite{8626048,shirani2022thompson}.

In addition, a Gaussian process based model predictive control (GP-MPC) method is performed for the system \eqref{eq:sys} to compare with the proposed ALC method. The GP-MPC method is a data-driven approach that employs a Gaussian process to model the unknown dynamics and then uses the model predictive control method to optimize the control law. To learn the system dynamics, $100$ samples are randomly collected from the system \eqref{eq:sys}. Then, the Gaussian process is trained using exponential kernel function, and the hyper-parameters are optimized using marginal log-likelihood approach. The cost function for the GP-MPC is set to be 
\begin{align}
J_{\rm MPC}=\sum\limits_{k=1}^{N_{\rm MPC}} \left( x(k)^\top Q x(k) + u(k)^\top R u(k) \right),
\end{align} 
with $N_{\rm MPC}=8$ as the prediction horizon, and $Q$ and $R$ are set the same as in the aforementioned ALC. 

Moreover, to demonstrate the influence of the initial control law on the training results, we present results for two ALC methods initialized with different control laws. Specifically, ${\rm ALC}_1$ uses the initial control law $\check{\bm g}({\bm x})=x_1+x_1^2+x_2^2+3x_1x_2$, while ${\rm ALC}_2$ uses $\check{\bm g}({\bm x})=\sin(x_1)-2x_2^2+2x_1x_2+1$. The comparison results are provided in Table~\ref{tab:comparison}. In the table, ``Reward'' denotes the closed-loop control reward for a segment of length 50, ``Learning Time'' indicates the time consumed by the learning process, and ``Simulation Time'' refers to the simulation time for a single segment (unit: seconds).
In the GP-MPC method, the learning process refers to the data collection and Gaussian process model training, while in the ALC method, the learning process refers to running the TS algorithm for 200 segments. The Reward and Simulation Time shown for the ALC method correspond to the closed-loop control reward and simulation time at the 200th segment.

It can be seen from Table~\ref{tab:comparison} that the reward obtained by the ALC method is affected by the choice of initial control law. A generally better choice is to select an initial control law ${\bm g}(\cdot)$ that has a similar structure to the dynamic equation $f(\cdot)$, as in ${\rm ALC}_2$, which enables the learned control law to achieve effects similar to feedback linearization, resulting in better control performance. In situations where the system dynamics are unknown, the selection of the initial control law may lead to a large reachable error $\Delta_{\mathcal{G}}$, but the upper bound of the reachable error is proved to be finite in Corollary \ref{cor:err}. On the other hand, the proposed ALC method only requires evaluating a deterministic parametric function when computing the control law, which results in much lower computational complexity and, consequently, significantly reduced simulation time per segment compared to the GP-MPC method.

\begin{table}[htbp]
	\centering
	\caption{Comparison of the proposed ALC method and the GP-MPC method.}
	\label{tab:comparison}
	\begin{tabular}{cccc}
		\toprule
		Method & Reward & Learning Time & Simulation Time \\
		\midrule
		GP-MPC & 0.401 & 1.543 & 7.853\\
		${\rm ALC}_1$ & 0.127 & 8.966 & 0.0016 \\
		${\rm ALC}_2$ & 0.396 & 9.815 & 0.0015 \\
		\bottomrule
	\end{tabular}
\end{table}

\begin{figure}[htbp]
	\centering
	\includegraphics[width=0.85\linewidth]{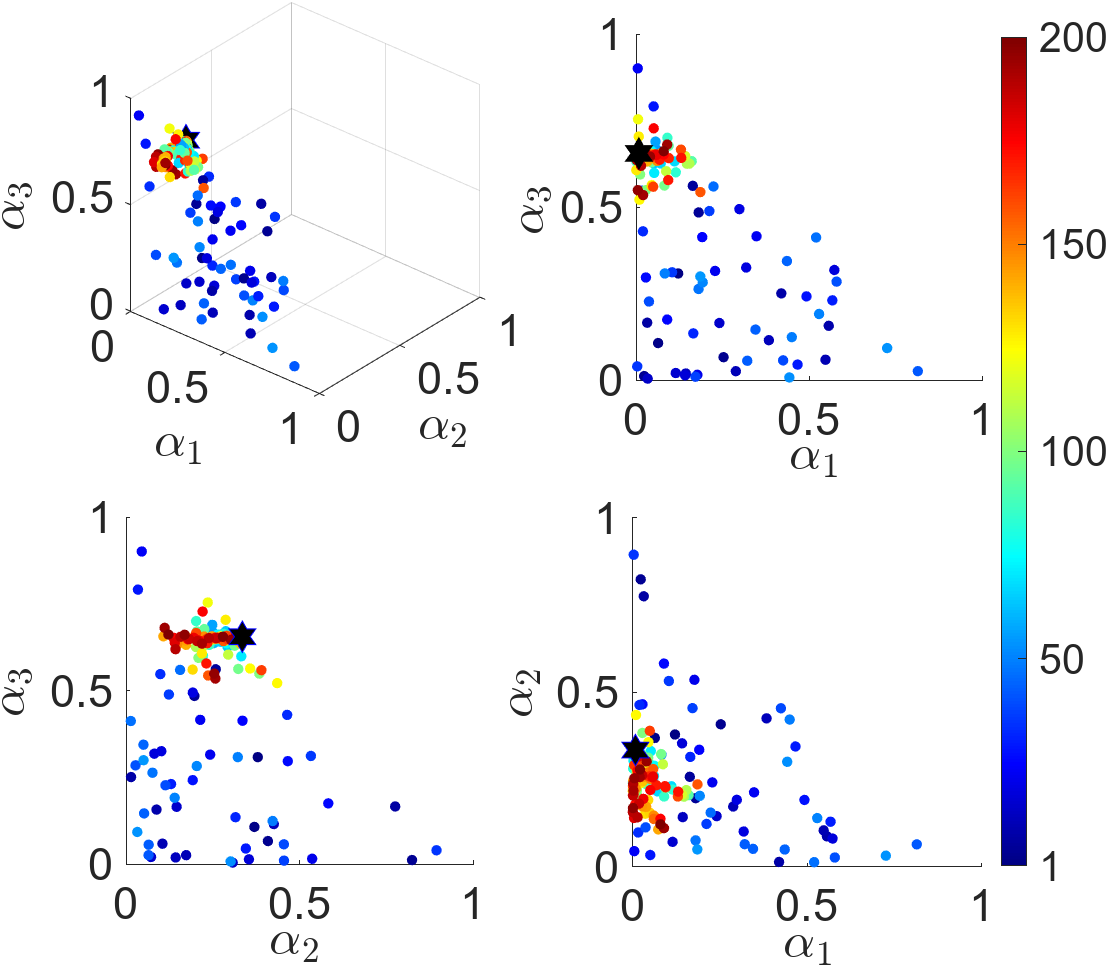}
\caption{The convergence process in a single test. The black hexagram represents the control law with maximized reward (namely, $\bar{\bm g}$), and the solid dots represent individual samples. The color of each dot indicates the sampling order: blue dots correspond to samples taken at the beginning of the learning process, green and yellow dots represent samples after some exploration, and red dots indicate samples from the 150th to the 200th segments.}
	\label{fig:sample_ALC}
\end{figure}

Furthermore, to demonstrate the convergence process of the control law, four different perspectives are displayed in the four subplots of Figure~\ref{fig:sample_ALC} for ${\rm ALC}_2$. As observed in the figure, as the proposed TS method progresses, the sampled control laws converge toward the control law $\bar{\bm g}$ which achieves the highest reward.

\section{Conclusion}\label{sec:Conclusion}
This work proposes an ALC method for unknown nonlinear systems based on the TS algorithm, aiming to optimize controller design through online learning. 
The proposed method constructs a Hilbert space $\mathcal{G}$ for potential candidate control laws and optimizes the selection of control laws within this space using online data. 
The parameterization method provides a well-structured function space for control law optimization, and the resulting potential performance degradation is proved to be bounded. Furthermore, through an active online sampling strategy, the learned cost function is proved to converge to a neighborhood of the true cost function exponentially. Additionally, the control regret during the online learning process is analyzed, including the constant regret caused by the function space construction and the exponentially convergent regret during the online learning process.
Numerical experiments validate the effectiveness of the proposed method and the correctness of the related theorems. Overall, the TS method effectively balances exploration and exploitation, rapidly improving control performance in the early stages of learning. However, the exploration near the steady-state may lead to a slight decrease in control performance. Future work will focus on further improving the convergence speed and control performance in the steady-state phase of the learning process.

\section*{Acknowledgment}
The authors would like to thank the Associate Editor and the anonymous reviewers for their suggestions which have improved the quality of the work. Moreover, The authors would like to express their gratitude to Yongjie Liu and Yuchen Yang from Peking University for their valuable assistance in improving the quality of the manuscript. 


%


\bibliographystyle{IEEEtran}
\bibliography{kk.bib}



\begin{IEEEbiography}[{\includegraphics[width=1in,height=1.25in,clip,keepaspectratio]{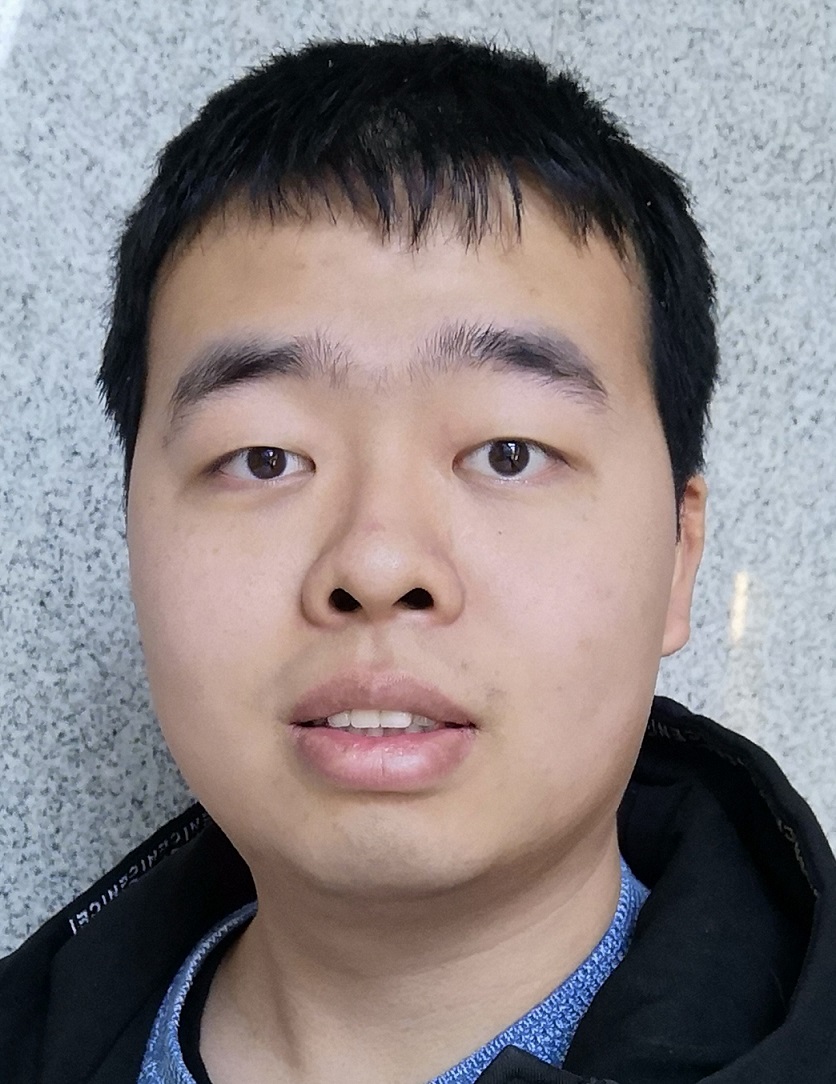}}]{Kaikai Zheng} was born in Xianyang, Shaanxi Province, China. He received the B.Eng. degree in automation from Beijing Institute of Technology, Beijing, China, in 2020. He is taking successive postgraduate and doctoral programs at the School of Automation, Beijing Institute of Technology. His research interests include event-based state estimation, system identification, event-triggered learning, and model predictive control.
\end{IEEEbiography}

\begin{IEEEbiography}[{\includegraphics[width=1in,height=1.25in,clip,keepaspectratio]{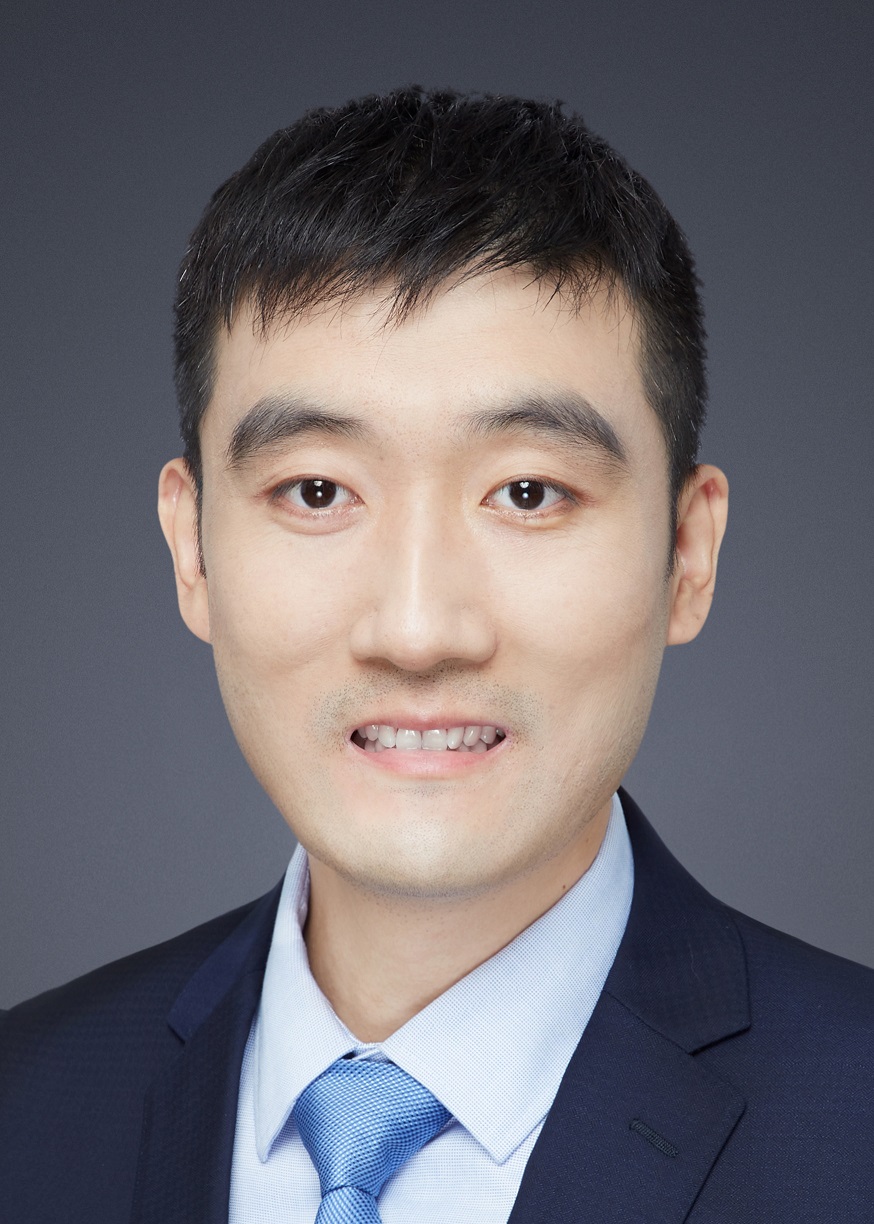}}]{Dawei Shi} received the B.Eng. degree in electrical engineering and its automation from the Beijing Institute of Technology, Beijing, China, in 2008, the Ph.D. degree in control systems from the University of Alberta, Edmonton, AB, Canada, in 2014. In December 2014, he was appointed as an Associate Professor at the School of Automation, Beijing Institute of Technology. From February 2017 to July 2018, he was with the Harvard John A. Paulson School of Engineering and Applied Sciences, Harvard University, as a Postdoctoral Fellow in bioengineering. Since July 2018, he has been with the School of Automation, Beijing Institute of Technology, where he is a professor. 
											
His research focuses on the analysis and synthesis of complex sampled-data control systems with applications to biomedical engineering, robotics, and motion systems. He serves as an Associate
Editor/Technical Editor for IEEE Transactions on Industrial Electronics, IEEE/ASME Transactions on Mechatronics, IEEE Control Systems Letters, and IET Control Theory and Applications. He is a member of the Early Career Advisory Board of Control Engineering Practice. He was a Guest Editor for European Journal of Control. He served as an associate editor for IFAC World Congress and is a member of the IEEE Control Systems Society Conference Editorial Board. He is a Senior Member of the IEEE.
\end{IEEEbiography}
										
										%

\begin{IEEEbiography}[{\includegraphics[width=1in,height=1.25in,clip,keepaspectratio]{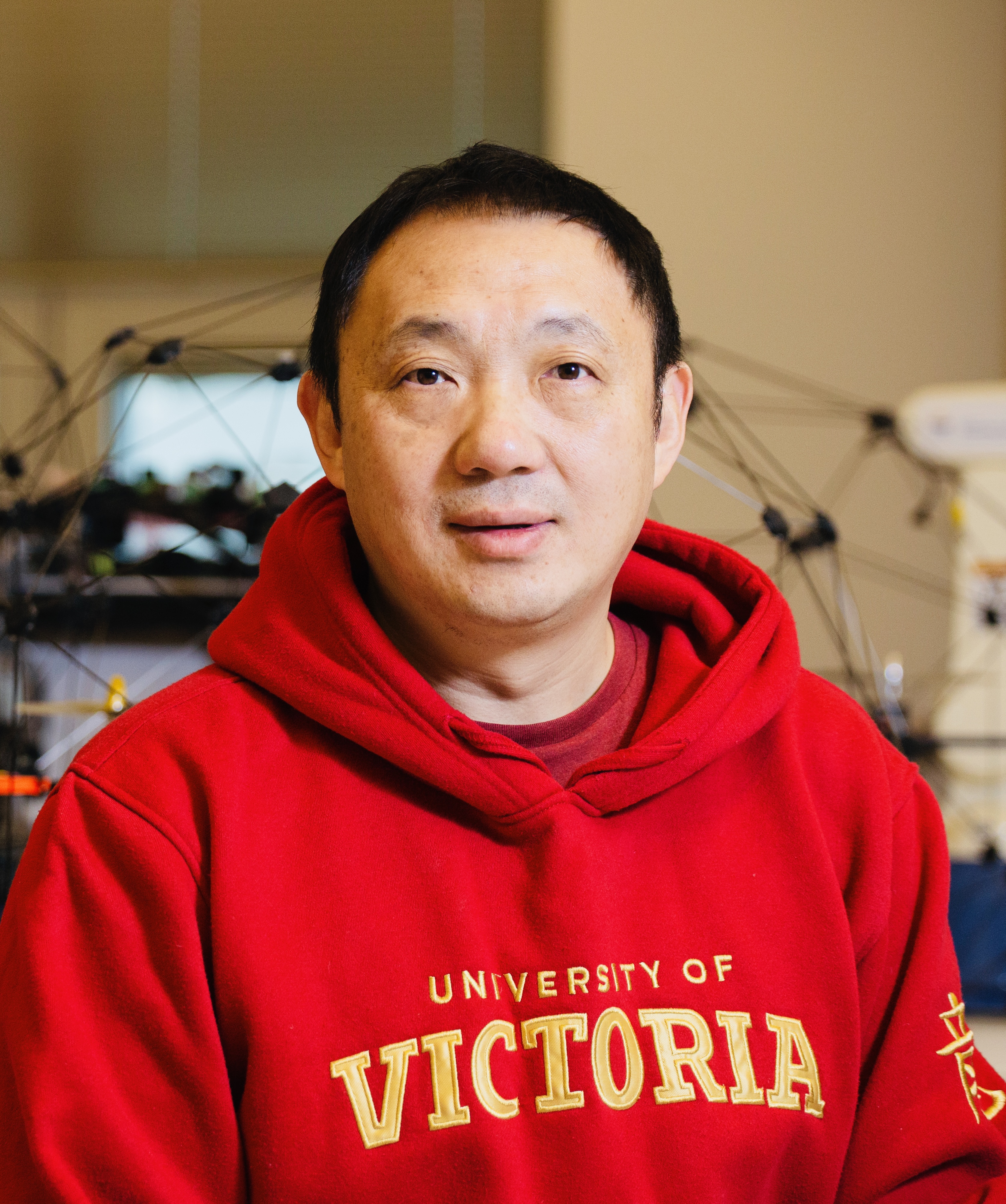}}]{Yang Shi} received the Ph.D. degree in electrical and computer engineering from the University of Alberta, Edmonton, AB, Canada, in 2005. From 2005 to 2009, he was an Assistant Professor and an Associate Professor in the Department of Mechanical Engineering, University of Saskatchewan, Saskatoon, SK, Canada. In 2009, he joined the University of Victoria, Victoria, BC, Canada, where he is currently a Professor in the Department of Mechanical Engineering. His current research interests include networked and distributed systems, model predictive control (MPC), cyber-physical systems (CPS), robotics and mechatronics, navigation and control of autonomous systems (AUV and UAV), and energy system applications.
											
Dr. Shi received the University of Saskatchewan Student Union Teaching Excellence Award in 2007, and the Faculty of Engineering Teaching Excellence Award in 2012 at the University of Victoria (UVic). He is the recipient of the JSPS Invitation Fellowship (short-term) in 2013, the UVic Craigdarroch Silver Medal for Excellence in Research in 2015, the 2016 IEEE Transactions on Fuzzy Systems Outstanding Paper Award, the Humboldt Research Fellowship for Experienced Researchers in 2018. He has served as the Vice-President on Conference Activities of IEEE IES, and Chair of IES Technical Committee of Industrial Cyber-Physical Systems. He is  the Editor-in-Chief for IEEE Transactions on Industrial Electronics; he also serves as Associate Editor for Automatica, IEEE Transactions on Automatic Control, IEEE Transactions on Cybernetics, etc. He is General Chair of the 2019 International Symposium on Industrial Electronics (ISIE) and the 2021 International Conference on Industrial Cyber-Physical Systems (ICPS). He is a Fellow of IEEE, ASME, Engineering Institute of Canada (EIC), and Canadian Society for Mechanical Engineering (CSME), and a registered Professional Engineer in British Columbia, Canada.
\end{IEEEbiography}
						
\begin{IEEEbiography}[{\includegraphics[width=1in,height=1.25in,clip,keepaspectratio]{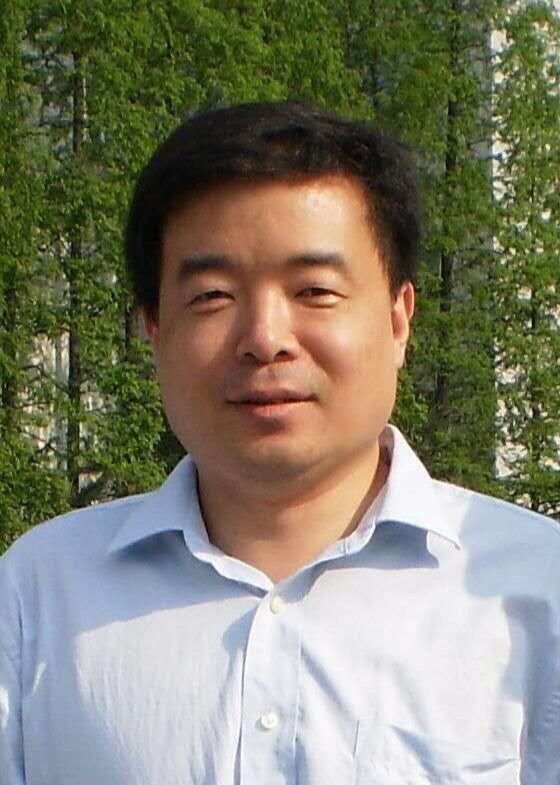}}]{Long Wang}  was born in Xi’an, China. He received the bachelor’s degree in automatic control from Tsinghua University, Beijing, China, in 1986, and the Ph.D. degree in dynamics and control from Peking University, Beijing, in 1992. From 1993 to 1997, he held postdoctoral research positions with the University of Toronto, Toronto, ON, Canada (with Prof. Bruce A. Francis) and the German Aerospace Center, Mu
	nich, Germany (with Prof. Juergen E. Ackermann). He is currently the Cheung-Kong Chair Professor of Dynamics and Control and the Director of the Center for Systems and Control, Peking University. He has supervised more than 50 Ph.D. students. He has also given a number of plenary lectures at major conferences, including Chinese Control Conference, Chinese Conference on Systems Science, and Chinese Conference on Intelligent Automation. His research interests include complex networked systems, evolutionary game dynamics, artificial intelligence, and biomimetic robotics.
	
Dr. Wang was the recipient of the National Science Prize (twice in 1999 and 2017), the Guan Zhaozhi Control Theory Award, the Zhang Siying Award in Decision and Control, and the Best Paper Awards for journal publications in Control Theory and pplications, and Science China Information Sciences.
\end{IEEEbiography}

\end{document}